\documentclass[journal, twoside]{IEEEtran}
\let\labelindent\relax
\usepackage{graphicx}
\usepackage{subcaption}
\usepackage{eqlist}
\usepackage{amsfonts}
\usepackage{tabularx}
\usepackage{booktabs}
\usepackage{amssymb}
\usepackage{amsmath}
\usepackage{enumitem}
\usepackage{threeparttable}
\usepackage[lined, ruled, linesnumbered, commentsnumbered]{algorithm2e}
\usepackage{lipsum}

\usepackage[usenames,dvipsnames]{xcolor}
\usepackage{comment}

\usepackage{multirow}

\begin{document}
\title{Multi-Pen Robust Robotic 3D Drawing Using Closed-Loop Planning}
\author{Ruishuang Liu$^{1}$, Weiwei Wan$^{1*}$, Keisuke Koyama$^{1}$ and Kensuke Harada$^{1,2}$
\thanks{$^{1}$Graduate School of Engineering Science,
        Osaka University, Japan}%
\thanks{$^{2}$National Inst. of AIST, Japan. 
        *Correspondent author: Weiwei Wan:
        {\tt\small wan@sys.es.osaka-u.ac.jp}}%
}

\markboth{ARXIV Preprint. Under Review for Formal Publication, 2020}
{Liu \MakeLowercase{\textit{et al.}}: Multi-Pen Robust Robotic 3D Drawing using Closed-loop Planning}
\maketitle

\begin{abstract}
This paper develops a flexible and robust robotic system for autonomous drawing on 3D surfaces. The system takes 2D drawing strokes and a 3D target surface (mesh or point clouds) as input. It maps the 2D strokes onto the 3D surface and generates a robot motion to draw the mapped strokes using visual recognition, grasp pose reasoning, and motion planning. The system is flexible compared to conventional robotic drawing systems as we do not fix drawing tools to the end of a robot arm. Instead, a robot selects drawing tools using a vision system and holds drawing tools for painting using its hand. Meanwhile, with the flexibility, the system has high robustness thanks to the following crafts: First, a high-quality mapping method is developed to minimize deformation in the strokes. Second, visual detection is used to re-estimate the drawing tool's pose before executing each drawing motion. Third, force control is employed to avoid noisy visual detection and calibration, and ensure a firm touch between the pen tip and a target surface. Fourth, error detection and recovery are implemented to deal with unexpected problems. The planning and executions are performed in a closed-loop manner until the strokes are successfully drawn. We evaluate the system and analyze the necessity of the various crafts using different real-word tasks. The results show that the proposed system is flexible and robust to generate a robot motion from picking and placing the pens to successfully drawing 3D strokes on given surfaces.
\end{abstract}

\section{Introduction}
\IEEEPARstart{T}{his} paper develops a flexible and robust robotic system for drawing on 3D surfaces using visual detection, planning, and control. We especially pay our attention to the drawing motion's autonomous planning considering the following three aspects: 3D surface, flexibility, and robustness. First, the system accepts 2D drawing strokes and a 3D mesh model or 3D point cloud of the target surface as input. It maps the 2D strokes to the 3D surface for robotic motion trajectory planning. Second, the drawing pens are not fixed to the flange of a robot arm. Instead, the robot plans grasp poses to pick up a drawing pen and holds the pen using its hand to draw the mapped 3D strokes. The robot is able to change the pens flexibly to draw with different colors. Third, since detecting, grasping, and manipulating pens may lead to errors, the system leverages in-hand pen pose estimation, force control, and error detection and recovery to correct the errors and ensure robust executions. The system works in a closed-loop manner until the strokes are successfully drawn.

\begin{figure}
    \centering
    \includegraphics[width=.99\linewidth]{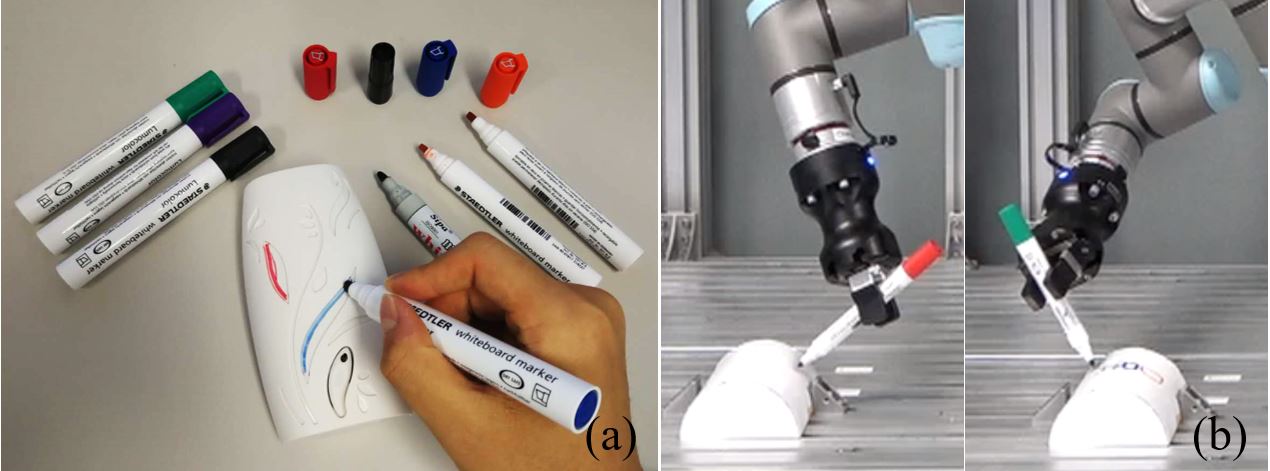}
    \caption{(a) A human draws on a 3D surface by changing multiple pens for different stroke colors. (b) This paper aims to develop a robust system that performs similar multi-pen drawing on 3D surfaces using closed-loop planning, considering visual and force feedback.}
    \label{figteaser}
\end{figure}

Robotic drawing can be viewed as moving a pen along a trajectory in a certain pattern concerning the target surface. Previously at manufacturing sites, the trajectories were taught in a point-to-point way by professional system integration engineers. The process was time-consuming and required expert knowledge. Modern researchers and developers solve the problem using closed-loop planning. For example, \cite{conner2005paint}\cite{chen2008automated}\cite{from2010optimal}\cite{andulkar2015novel} developed algorithms for painting path planning and optimization on a free-form surface to ensure evenly distribution of the paint. Painting is a simplified robotic drawing case as there is no contact between a robotic end-effector and a painting surface. In contrast, robotic drawing, especially drawing with hand-held pens, is a contact-rich manipulation problem. Closed-loop planning using vision and force feedback is an important solution to maintaining contact between the pen and the target painting surface and is widely studied \cite{kudoh2009painting}. Control methods like impedance control play an important role in helping to achieve a stable drawing motion \cite{song2018artistic}. Although we can find lots of state-of-the-art studies that close the loop and realize robust and practical robotic systems, they have an assumption that the drawing tools are fixed to the end flange of a robot arm, which significantly influences the flexibility in switching among diverse stroke colors and stroke types.

In the human world, artists leverage rich tactile and force feelings to control a drawing pen and frequently change their pens during the drawing to switch to different stroke colors or stroke types, as is shown in Fig.\ref{figteaser}(a). Inspired by human artists, we in this paper develop a robotic system that performs similar multi-pen drawing on 3D surfaces using closed-loop planning, considering visual and force feedback. The system's planning component is based on our previous work, which enables a robot to use tools with a general two-finger gripper \cite{chen2019combined}. We improve the planner to realize a flexible and robust robotic 3D drawing system by including 3D stroke mapping and vision-force feedback. We assume the multiple pens and grippers are directly used without any modification. There is no need for human intervention -- The robot autonomously reasons grasp poses, regrasp poses, and plans joint motion following the mapped 3D goal strokes while considering vision and force feedback.

In the remaining part of this paper, we present the technical details of the developed system. We highlight our contributions in the following aspects. (1) We do not fix drawing tools to the end of a robot arm. Instead, we allow a robot to select drawing pens using a vision system and hold the pen to draw a general two-finger gripper. (2) We develop a metrology based high-quality mapping method to map pre-specified 2D goal strokes to 3D target surfaces without deformation. (3) We develop an in-hand pose estimation method to correct accumulated errors and trigger a re-planning if no solution was found. (4) We employ force control to avoid noisy visual detection and calibration, thus ensuring a firm touch between the pen tip and a target surface. (5) We implement error detection and recovery to deal with other unexpected problems. Our system's planning and executions are performed in a closed-loop manner until the strokes are successfully drawn. In the experimental section, we evaluate our system using various real-world tasks. The results show that the proposed system is flexible and robust to generate a robot motion from grasping the pens to finishing the drawing.

The paper's organization is as follows: First, we review the work related to our research in Section II. Then, we outline an overview of the system architecture in Section III. Detailed descriptions of the mapping and feedback planning are presented respectively in Section IV and V. Experiments, comparisons, and analyses are carried out in Section V. Conclusions and discussions on the future work are presented in the last section.

\section{Related Work}
We review the related studies of our work in three aspects: Robotic drawing system; Mapping 2D strokes to 3D object surfaces; Reasoning and motion planning considering the usage of tools.

\subsection{Robotic Drawing Systems}
One of the historical work that uses machines to draw pictures is \cite{malina1991aaron}. The book presented a computer program named Aaron, which is able to draw figures and shapes on paper using a pen driven by a mechanical plotting machine. Since then, developing automated systems to draw pictures is continuously an important and popular topic in robotics and automation science communities. With the rapid development of robotic hardware and computer vision, many researchers studied drawing using multi-joint robots. The diversity and flexibility of automated drawing systems increased significantly with the help of multi-joint robots. For example, Calinon et al. developed a portrait drawing system using a 4-DoFs robotic arm \cite{calinon2005humanoid}. Tresset et al. \cite{tresset2013portrait} presented a more advanced portrait drawing robot named ``Paul'', which was able to draw portraits using the equivalent of an artist's stylistic signature based on several processes that mimic drawing primitives or skills. Jun et al. \cite{jun2016humanoid} used a humanoid robot to draw a large picture on a wall. Sasaki et al. \cite{sasaki2016visual} developed a deep learning method to learn drawing motion with a given target image directly. These robotic drawing systems use ordinary pens to draw on a 2D plane. Different from them, other researchers explored using more diverse drawing tools like brushes. Diverse drawing tools made the motion planning and control of the robotic drawing system challenging and interesting. For example, Yao et al. \cite{yao2005painting} proposed an approach to deformable brush control and developed a painting robot that was able to draw bamboos by using a writing brush. Wang et al. \cite{wang2020robot} developed a Chinese calligraphy robot using a dynamic model of a writing brush. Drawing on non-planar surfaces is also a popular topic and received much research interest. For example, Lam et al. \cite{lam2007robot} developed an automated 3-DoFs sketching system to conduct pen drawing on a 2.5D surface. Song et al. \cite{song2018artistic} realized pen drawing on an arbitrary surface with force feedback.

Compared to the aforementioned robotic drawing systems, our difference is we focus on multi-pen drawing on 3D surfaces. We develop computational geometric algorithms to map 2D strokes to 3D surfaces or 3d point clouds, develop reasoning and motion planning algorithms to select and change pens, and close the planning loop using vision and feedback control.

\subsection{Mapping 2D Strokes to 3D Object Surfaces}

An intuitive solution to 2D-to-3D mapping used in the robotic drawing systems is to project 2D strokes onto $xy$-plane and estimate $z$-value according to the surface's shape. This idea is used in \cite{song2018artistic}\cite{lam2007robot}. Although this method could successfully create 3D strokes, they suffer from deformation problems. The Euclidean distances of the mapped 3D strokes get distorted from its 2D origin. Thus, researchers explore deformation-free mapping, which is essentially a surface parameterization problem in computer graphics \cite{floater2005surface}\cite{zheng20082d}. One effective method for surface parameterization is using Least Squares Conformable Mapping (LSCM) \cite{haker2000conformal}\cite{levy2002least}. The method is used in computer graphics for creating a UV map from mesh models to textures. Song et al. \cite{song2019distortion} leveraged the LSCM method to implement distortion-free stroke mapping. They spread 3D mesh models onto 2D planes using LSCM, mapped 2D strokes on the spreading plane, and wrapped back to create 3D strokes. 

Some other deformation-free mapping methods do not parameterize surfaces explicitly. Instead, they control the deformation in distances by using surface mapping metrology \cite{murphy2003stitching}. The methods are widely used in applications like mesh surface following. For example, Carmelo et al. \cite{mineo2016robotic} proposed a method for computing the mesh-following trajectories of a mesh surface by measuring distances along the intersection lines between the surface and a bunch of scanning planes. A dual-arm inspection system was developed based on the proposed method. Can et al. \cite{can2010five} presented an algorithm to project 2D patterns onto B-spline surfaces using similar metrology.

In our mapping, we both implement explicit parameterization based methods and develop metrology based methods. We compare the performance of these methods and use the high-quality one to generate 3D strokes.

\subsection{Reasoning and Motion Planning Considering Using Tools}
Three levels of planning are needed to enable the robot to manipulate multiple pens with a general gripper. The first level is grasp pose reasoning, where the goal is to find common grasp poses for starting and goal object poses. The reasoning problem has been studied a lot in robotic manipulation previously. For example, Saut et al. \cite{saut2010planning} developed a grasp planner to plan grasp poses for multi-finger hands, and then used the planned grasp poses to reason pick-and-place sequences. Wan et al. \cite{wan2015improving} developed similar approaches and analyzed its performance for assembly tasks.

After finding common grasps poses, the next step is to select a proper pair from all candidates and plan the motion between the selected pair. The related problem is grasp optimization\cite{liu2018grasp} and constrained motion planning and control\cite{mirabel2016}. Grasp optimization relies on the chosen quality measures. Previously, several different qualities have been developed and used in optimization \cite{roa2015grasp}\cite{zheng2019computing}\cite{tsuji2009easy}. Traditional motion planning algorithms include the probabilistic roadmaps approaches to search collision-free motion in the joint space \cite{simeon2004manipulation}\cite{kavraki1996probabilistic}, and the Rapidly-exploring Random Trees (RRT) \cite{lavalle2000rapidly}. Incorporating task constraints into the planning process represents significant challenges \cite{kingston2018sampling}. Toussaint et al. studied the tool-use planning in domains that include physical interactions \cite{toussaint2018differentiable}. Rachel et al. proposed a method to formulate the force constraints in the tool manipulation tasks\cite{holladay2019force}.

Previously in our group, we also developed reasoning, selection, and constrained planners for robots to manipulate tools. Our studies were under the combined task and motion planning framework \cite{wolfe2010combined} \cite{srivastava2014combined}. For example, Raessa et al. \cite{raessa2019teaching} presented a method to teach a dual-arm robot to use common electric tools. Chen et al. \cite{chen2019combined} designed a motion planner to manipulate a suction cup tool. 

In this work, we use similar combined task and motion planning methods to reason the grasp poses for multiple pens, and generate manipulation and drawing motion considering pen contact constraints. We particularly focus on using in-hand pose estimation during the combined planning to make the system robust and recoverable from large deviations and errors.

\section{Overview of the Workflow}

\begin{figure}
    \centering
    \includegraphics[width=.95\linewidth]{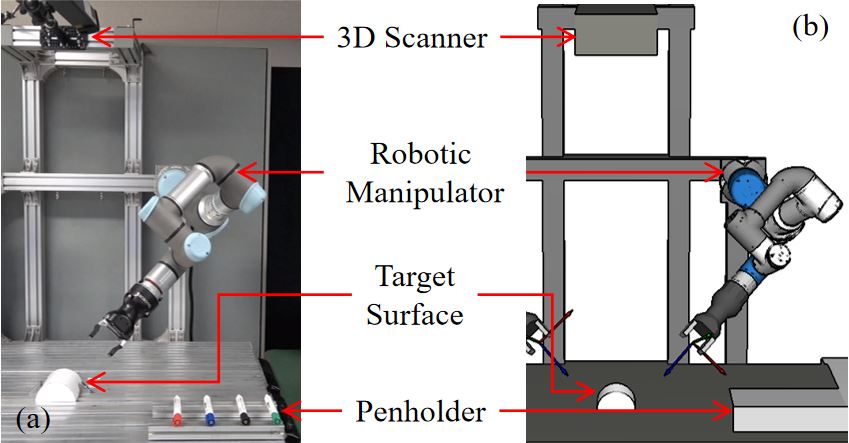}
    \caption{One exemplary hardware setup for the robotic 3D drawing system. (a) A picture of the hardware. (b) The simulation environment.}
    \label{fig:setup}
\end{figure}

This section presents an overview of the methods and workflow to give readers an intuitive idea of our system. First, we show one hardware setup and its corresponding planning interface in Fig.\ref{fig:setup}. Note that the proposed method is not limited to this hardware setup. We present it here to give readers a solid conception. A robotic manipulator with a general 2-finger gripper is used to manipulate pens and draw pictures in the setup. A 3D depth sensor is installed on top of the workspace to detect pens and target surfaces. The pens and surfaces are placed randomly (with adequate clearance) on the table under the vision sensor to wait for robotic execution.

\begin{figure}[!htbp]
    \centering
    \includegraphics[width=.95\linewidth]{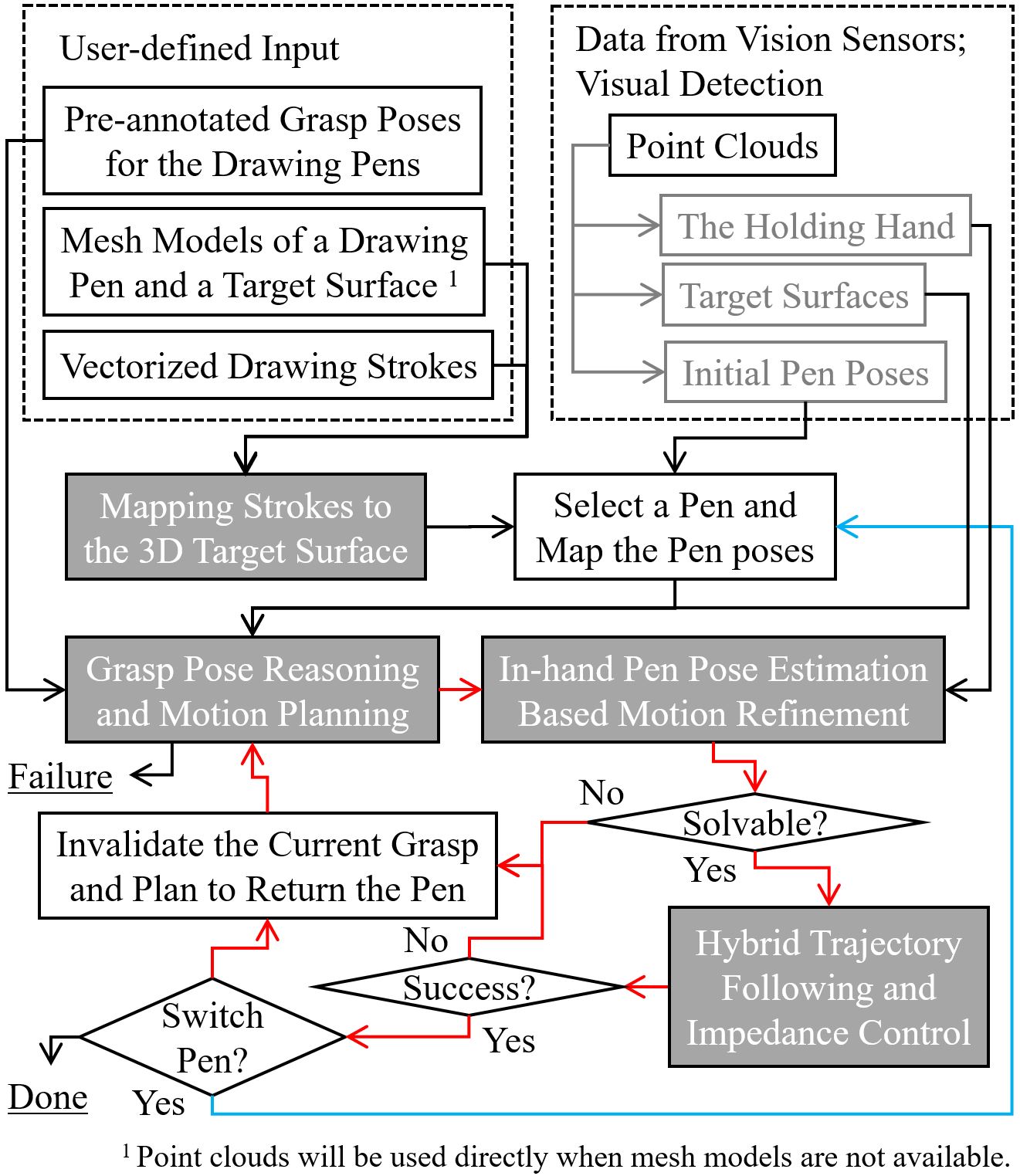}
    \caption{The workflow of the proposed robotic 3D drawing system. The two dashed boxes are external input. The gray boxes indicate the essential algorithms. The red arrows illustrate the closed-loop planning. The gray boxes show the planned motion details and how they are connected to the in-hand pose estimation to close the loop.}
    \label{fig:workflow}
\end{figure}

A diagram of our workflow concerning the shown hardware setup is shown in Fig.\ref{fig:workflow}. The workflow starts from two inputs shown by the dashed boxes in the figure. The first input is from users. It includes the pre-annotated or pre-planned grasp poses for the drawing pens, mesh models of the pens, and vectorized drawing strokes on a 2D surface. The second input is the point clouds obtained using a 3D vision sensor. The points clouds are divided into two areas where the system finds the initial pen poses in the first area and finds the target surface to draw the strokes in the second area.

The four gray boxes in the lower part of the workflow indicate the essential algorithms. The ``Mapping Strokes to the 3D Target Surface' box maps the vectorized 2D strokes to the 3D target surface detected using the point clouds from the second input. It accepts mesh models and the vectorized 2D strokes from the first input. Together with the initial pen poses detected from the second input, it will produce a sequence of kinematic pen poses. The pose pen sequence will be sent to the second gray box named ``Grasp Pose Reasoning and Motion Planning'', to determine grasp poses and plan manipulation motion. The reasoner will iterate throughout all available grasp poses in a database to find the candidates that can finish all the motion. The planned motion includes the motion to pick up a pen, move a pen to a visible position for in-hand pose estimation, and draw the mapped strokes. The motion to move to a visible position for in-hand pose estimation and the drawing motion will be regulated by the third gray box, namely the ``In-hand Pen Pose Estimation and Re-planning'' box. The box determines if there is an in-hand error and will trigger a refinement or re-planning according to the errors. Finally, the ``Hybrid Trajectory Following and Impedance control'' box ensures a firm contact between the pen tip and the target surface. It performs error detection and triggers error recovery in an emergency.

The planning process has a closed-loop, as illustrated by the red arrows in the diagram. If a refined pose or re-planning is not solvable, the system will invalidate the current grasp pose, iterate to the next grasp pose from the reasoned candidate set, and repeat the reasoning and planning. The reasoner and motion planner will scan all candidate grasp poses until it finds solutions for all strokes or reports a failure. Also, during execution, the robot leverages impedance control to make sure a stroke is clearly drawn on the target surface. It uses a hybrid trajectory following to monitor the position deviation. If there is a larger deviation, the system will trigger an error recovery process. In case of an unrecoverable error, the workflow will go back to the ``In-hand Pen Pose Estimation'' box to re-estimate the pen's pose and re-generate the motion by continuing from the failure point. The details of the four essential algorithms and the closed-loop planning will be presented and examined in the remaining sections.

\section{Mapping Strokes to 3D Surfaces}
The goal of ``Mapping Strokes to the 3D Target Surface'' is to map a sequence of vectorized drawing strokes (can be considered as points) in $\mathbb{R}^2$ to points on the target drawing surface in $\mathbb{R}^3$ while preserving the relationship between the points. We expect an ideal mapping method to comprise the following properties: (1) The method can cope with any target surface, both raw point clouds or Computer-Aided Design (CAD) software-generated mesh models; (2) The method has minimal distortion in shape; (3) The method returns smooth drawing strokes that are easy to follow by robot arms. Following this consideration, we propose both metrology based methods and parameterization based methods in the following two subsections.

\subsection{Metrology Based Methods}
\begin{figure}[!htbp]
    \centering
    \includegraphics[width=.95\linewidth]{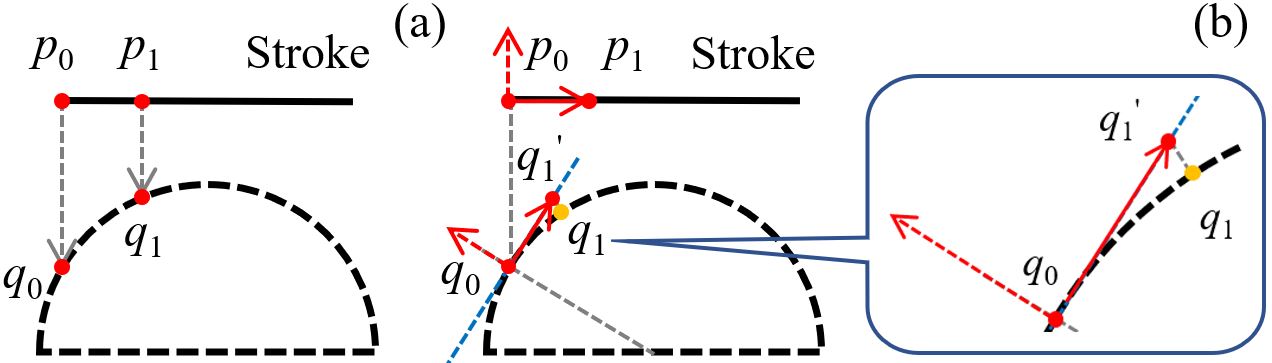}
    \caption{Projecting a stroke (the solid black segment at the top) to a curved surface (the dashed semicircle at the bottom). (a) The intuitive mapping method. (b) The proposed method.}
    \label{fig:projection}
\end{figure}

To better explain the metrology based methods, we first present an intuitive mapping method, namely the orthogonal projection mapping, as shown in Fig.\ref{fig:projection}(a). In this case, the next point $p_1$ in the 2D stroke is directly mapped onto a target surface by projecting along the vertical direction $\overrightarrow{p_0q_0}$ (or $\overrightarrow{p_1q_1}$, they are the same). Since all points are projected in the same direction without considering surface normals, this intuitive method's distortion changes along with the target surface's curvature. In the figure, $\mathtt{len}(\overrightarrow{p_0p_1})>>\mathtt{len}(\overrightarrow{q_0q_1})$. It also does not work if the angle between the normal of a target surface and the projection direction is larger than 90 degrees. 

We revise the intuitive mapping method by considering the cross product of the target surface normals and the vectors (moving directions) of the strokes and propose the metrology based methods. Fig.\ref{fig:projection}(b) illustrates the fundamental idea. The mapping of the point $p_1$ is considered as a projection of $\overrightarrow{p_0p_1}$ to a tangential plane at $q_0$. The blue dash line in the figure illustrates the tangential plane. The dashed red arrows at $p_0$ and $q_0$ in the figure show the vertical direction perpendicular to a 2D stroke and the normal of the target surface at $q_0$, respectively. The solid red arrows between $p_0$ and $p_1$, and $q_0$ and $q_1^{'}$, show the vectors before and after projection. The final projected point is $q_1$ instead of $q_1^{'}$. It is further snapped to the mesh surface, as shown by the zoom box in Fig.\ref{fig:projection}(b). The projection and snapping together maintains the metrology -- The length of $\overrightarrow{p_0p_1}$ is the same as $\overrightarrow{q_0q_1^{'}}$. It is approximated by $\overrightarrow{q_0q_1}$ to minimize distortion.

The above process is formulated as the pseudo-code shown in Algorithm \ref{alg:projection}. The algorithm's input includes the target surface, a set of points in a stroke, and a starting point on the target surface. As mentioned at the beginning of this section, we expect the algorithm to apply to both raw point clouds or mesh surfaces. Thus, line 8 of the algorithm, say finding the nearest point to $q_{i+1}^{'}$ on $\mathcal{M}$, matters a lot. We achieve the expected performance by sampling the target surfaces in the case of a mesh or directly use the points in the case of point clouds. Then, we find the nearest point to $q_{i+1}^{'}$ from the samples. This implementation is direct and fast. Thus, we name it the Direct Implementation of the metrology based method (DI). A likely drawback of this method is the random samples or the noisy points from the point clouds may induce errors and lead to accumulated distortion and non-smooth strokes. To avoid the problem, we also proposed an Estimated Implementation of the metrology based method (EI), where we find a set of nearest points to $q_{i+1}^{'}$ from the samples, estimate a plane using the set, and compute ${q_{i+1}}$ by projecting $q_{i+1}^{'}$. Stroked mapped by the EI are expected to have a better quality than the 
DI. We will examine them later in the experimental section.

\begin{algorithm}[!htbp]
    \DontPrintSemicolon
    \SetKwInput{KwInput}{Input}                
    \SetKwInput{KwOutput}{Output}              
    \KwInput{ 
        $\mathcal{M}$, the target surface;\newline
        $\mathcal{P}$, a set of points in a stroke in $\mathbb{R}^2$;\newline
        $p_0 \in \mathcal{P}$, a start point on $\mathcal{P}$;\newline
        $q_0, n_0 \in \mathcal{M}$, a start point and its normal on $\mathcal{M}$;}
    \KwOutput{
        $\mathcal{R}$, a set of  points on $\mathcal{M}$;}
    \Begin{
        Scale $\mathcal{P}$ to fit the 2D work-space of $\mathcal{M}$;\\
        Convert $\mathcal{P}$ to $\mathbb{R}^3$ by adding a $z$ value with 0;\\
        $p_i\leftarrow p_0$, $n_i\leftarrow n_0$;\\
        \For{$point$ $p_{i+1}$ $\in \mathcal{P}$ }{
            $T \leftarrow$ {Rotation matrix from $(0,0,1)$ to $n_i$};\\
            $q_{i+1}^{'} \leftarrow$ {$p_i + T \times {\overrightarrow{p_ip_{i+1}}}$};\\
            $q_{i+1},n_{i+1} \leftarrow$ Nearest point to $q_{i+1}^{'}$ on $\mathcal{M}$;\\
            $p_i \leftarrow$ {$p_{i+1}$};\\
            $n_i \leftarrow$ {$n_{i+1}$};\\
        }
        \Return $\mathcal{R}$;
    }
    \caption{Drawing path rendering} 
    \label{alg:projection} 
\end{algorithm}

Algorithm \ref{alg:projection} is the routine for a single stroke. When there are multiple strokes, a linear interpolation is applied between the endpoint of a previous stroke to the start point of its next one. The same algorithm founds the projected position of the next stroke's start point. The points on the interpolated segment are ignored. 

\subsection{Parameterization Based Methods}

The parameterization based methods consider the projection problem inversely. Instead of directly projecting a 2D stroke onto a 3D surface, the parameterization based methods unfold the 3D surface onto a 2D plane, find the correspondence between a 2D stroke and the unfolded surface, and pack the correspondence back into 3D space. In this work, we carry out parameterization using the LSCM method \cite{levy2002least}. LSCM enables transforming a 3D graph into a 2D one while minimizing angles' deformations (not necessarily lengths). The LSCM parameterization returns the minimal-deformation 2D positions of all 3D vertices on the target surface.

The workflow of our parameterization based methods is as follows. First, we unfold mesh vertices or the point clouds to a 2D plane using LSCM. Then, we use a given starting position to find the correspondence between a 2D stroke and the unfolded 2D positions and pack the found correspondences back to 3D to obtain a 3D stroke. Throughout the process, the ``find the correspondence between a 2D stroke and the unfolded 2D positions'' step 
is critical. We propose two implementations to find the correspondence. In the first implementation, we find the nearest unfolded 2D position to a point in a 2D stroke, and only pack back this single point. We call it the Single-point Implementation of the parameterization method (SI). The implementation could be problematic as the uneven unfolding, or the point clouds' noisy points may induce errors and lead to non-smooth strokes. In the second implementation, we sample the target surface to find the k-nearest points of a point in a 2D stroke from the unfolded 2D positions, pack them back to 3D samples on the 3D surface, and perform interpolation to find the target point. This implementation is similar to the one introduced in \cite{song2019distortion}. We call this implementation the Interpolation Implementation of the parameterization based method (II). The performance of these two implementations will also be examined later in the experimental section. The most high-quality one of the four implementations will be used to generate 3D strokes.

\section{Reasoning and Closed-Loop Planning}
This section presents the remaining three essential algorithms -- the ``Grasp Pose Reasoning and Motion Planning'', the ``In-hand Pose Estimation'', and the ``Hybrid Trajectory Following and Impedance Control''. It comprises three subsections. In the first subsection, we present the methods used to detect the object poses. The remaining two sections explain the details of the three essential algorithms.

\subsection{Detecting the Object Poses}
A depth sensor is installed on top of the workspace to capture the objects' point clouds. The poses of the target surface and the pens are estimated by matching their mesh models to segmented sections of the captured point clouds. The segmentation and estimation could be performed easily by conventional algorithms like Density-Based Spatial Clustering of Applications with Noise (DBSCAN) based segmentation \cite{witten2002data}, RANdom SAmple Consensus (RANSAC) based global search \cite{zhou2016fast}, and Iterative Closest Point (ICP) based local refinement\cite{pomerleau2013comparing}.

\begin{figure}[!htbp]
    \centering
    \includegraphics[width=0.95\linewidth]{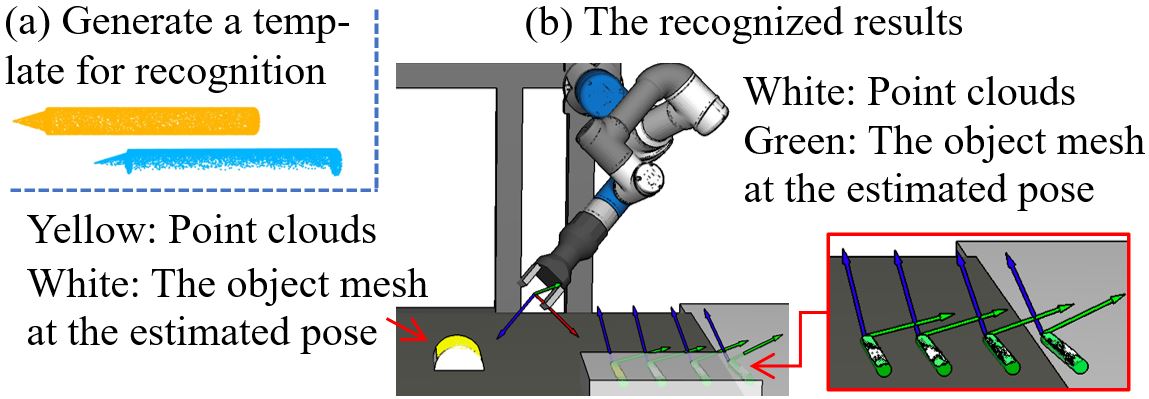}
    \caption{(a) Generate a partial view template for visual recognition. The surface of the partial view is sampled unevenly following the viewpoint direction to improve recognition performance. Orange: The full model of a drawing pen. Blue: An unevenly sampled partial view. (b) An example of detecting the initial target surface pose and pen poses. The meanings of the various colored illustrations are explained in the captions.}
    \label{fig:visual_detection}
\end{figure}

An important improvement we made to these conventional algorithms is that we sample the models unevenly considering the depth sensor's viewpoint. The old routine to perform the matching is: (1) Generate a partial view of the model; (2) Sample the partial view evenly to generate a template and compute the features; (3) RANSAC; (4) ICP. We revised step (2) to improve the precision of matching - Instead of sampling evenly, we leverage a changing sampling density considering the angles between the normals of the mesh surface and the vector from the mesh's geometric center to the position of the depth sensor. The sampling density is formulated as a function of the angle as follows:
\begin{equation}
    \rho = \frac{1}{1+\exp(\theta -\frac{\pi}{2})}-0.5
\end{equation}
where, $\rho$ is the symbol used to denotes the density of surface sampling. $\theta$ is the angle between the normal of a mesh triangle and the vector from the object's geometric center to the depth sensor's position. Under this equation's control, the triangles that face the camera will be sampled with higher density. The side triangles will be less sampled. The triangles on the back will be ignored. The uneven samples increase the fitness of RANSAC and ICP, thus improves the precision of pose estimation. Fig.\ref{fig:visual_detection} exemplifies the template created using uneven sampling, the captured point clouds, and the estimated surface pose and pen poses using the methods mentioned above. 

\subsection{Grasp Pose Reasoning and Motion Planning}

\begin{figure}[!htbp]
    \centering
    \includegraphics[width=0.95\linewidth]{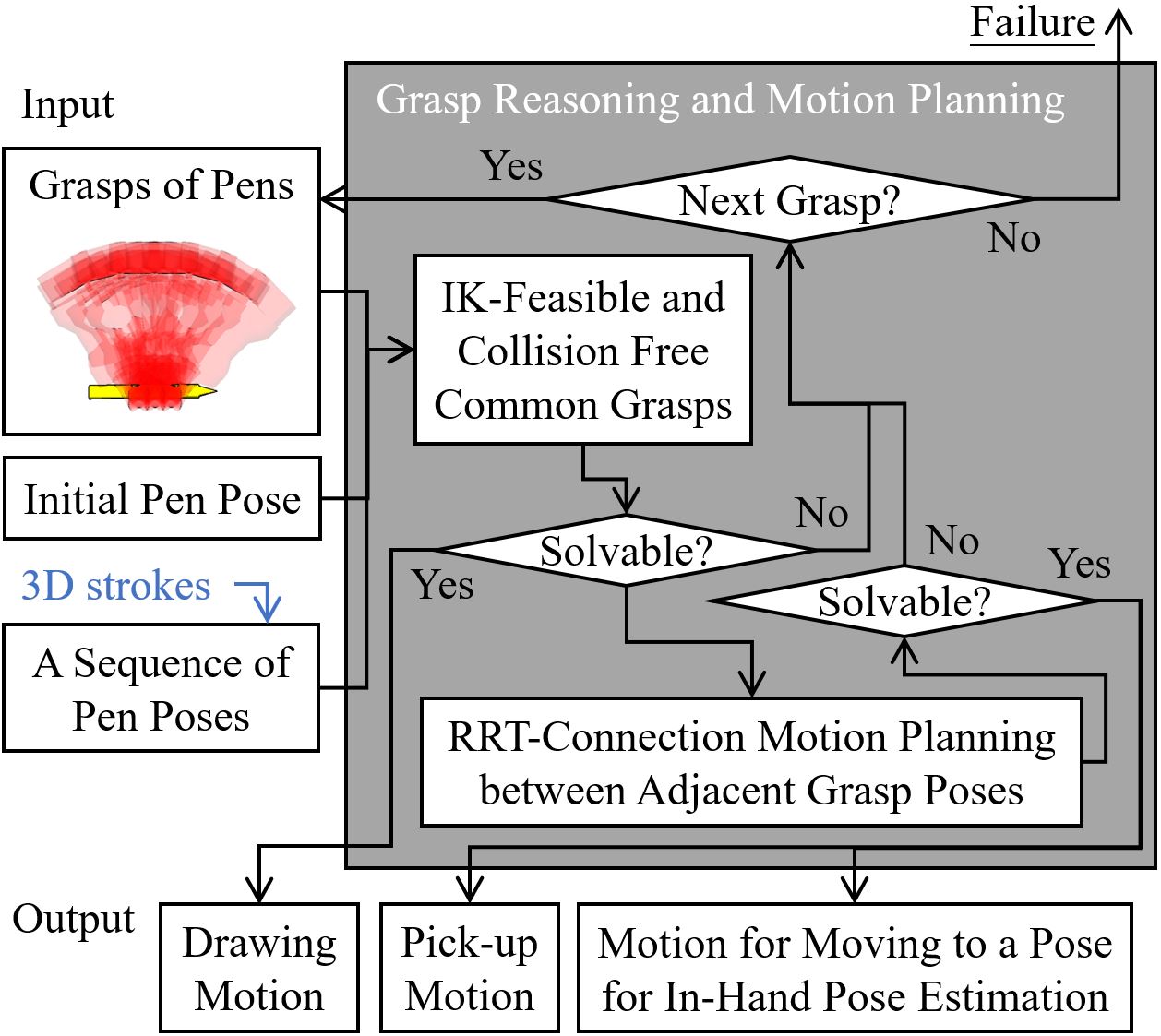}
    \caption{A detailed diagram of the ``Grasp Pose Reasoning and Motion Planning'' gray box in Fig.\ref{fig:workflow}.}
    \label{fig:motion_planner}
\end{figure}

The details of the ``Grasp Pose Reasoning and Motion Planning'' algorithm is shown by the diagram in Fig.\ref{fig:motion_planner}. This diagram is a direct expansion of the ``Grasp Pose Reasoning and Motion Planning'' gray box in Fig.\ref{fig:workflow}. It accepts (1) Pre-annotated grasp poses for the pens; (2) The estimated initial pen poses; (3) A sequence of the pen poses generated by attaching the pen tips to the mapped 3D strokes along with surface normal directions; as input. These data will be used in the ``IK-Feasible and Collision Free Common Grasps'' box to reason the IK-feasible and collision-free common grasps. The ``RRT-Connection Motion Planning between Adjacent Grasp Poses'' box will plan joint motion for the robot arm to move between adjacent common grasps in a sequence. The algorithm will iterate through all grasps until it finds the ``Drawing Motion'', the ``Pick-up Motion'', and the ``Motion for Moving to a Pose for In-Hand Pose Estimation'', or reports a failure.

Especially for the third input, namely the sequence of pen poses, we attach pen tips to each 3D stroke point while considering the surface normals. The attachment may lead to discontinuity due to the sudden change of surface normals and, consequently, reduce the generated drawing motion's robustness. The sudden changes get worse when there is no exact CAD model, and the target surface is made of raw point clouds captured by depth sensors. To avoid the discontinuity, we perform quaternion Spherical Linear Interpolation (SLERP) \cite{mortenson1999} to interpolate pen poses in-between the directly attached pen pose sequence.

\subsection{Using Closed-Loop Planning to Improve Robustness}
\label{close_loop_planning}

In the first subsection of this section, we showed that we improve visual estimation performance using uneven sampling. The method improves the precision of visual detection. However, even if we are perfect in detection, the errors cannot be removed entirely. There remain errors caused by (1) displacements of the pen during its interaction with the gripper, (2) calibration of the vision sensors, and (3) the numerical discretization of CAD models. In this subsection, we focus on these remaining errors and present our solutions. We use in-hand pen pose estimation to avoid the errors caused by (1), and use impedance control to eliminate the errors caused by the other two problems. Both the in-hand pose estimation and impedance control will form a closed-loop in case of a failure. The system will invalidate the current grasp pose and start over. Besides, we compare the positional difference to detect errors and trigger an error recovery mechanism to correspond to other unexpected problems. 

\subsubsection{In-hand pose estimation} 
\begin{figure}[!htbp]
    \centering
    \includegraphics[width=.825\linewidth]{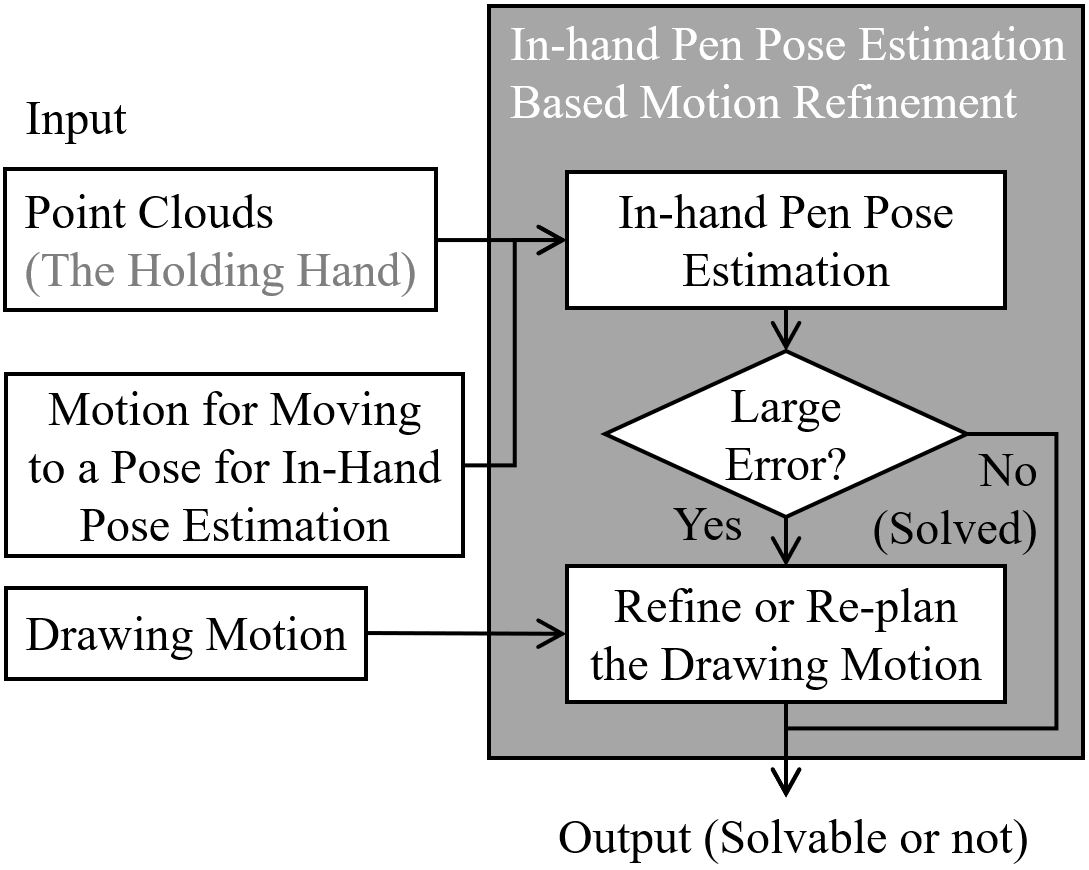}
    \caption{A detailed diagram of the ``In-hand Pose Estimation Based Motion Refinement'' gray box in Fig.\ref{fig:workflow}.}
    \label{figinhandflow}
\end{figure}

We use the same methods as detecting the initial object poses to perform in-hand pose estimation. The details of our algorithm are shown in Fig.\ref{figinhandflow}. We capture the point cloud near the robot gripper (its position could be obtained using joint encoder values and forward kinematics), estimate the pose of a grasped pen, and compare it to the ideal in-hand pose values used in the simulation. In case of a large error, we refine 
or re-plan the drawing motion to correct the trajectories.

\begin{figure}[!htbp]
    \centering
    \includegraphics[width=.95\linewidth]{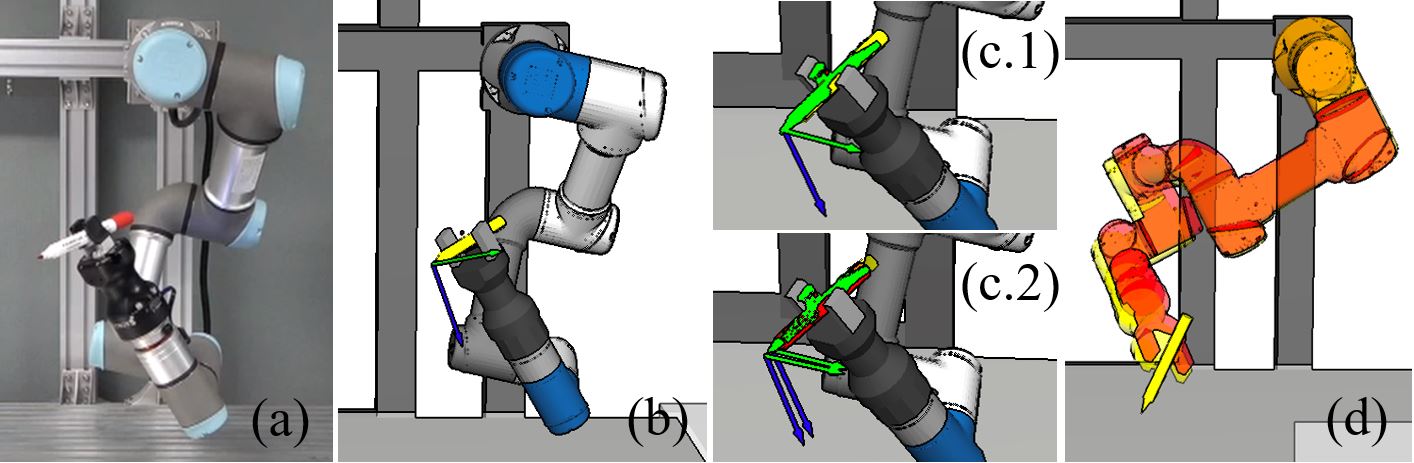}
    \caption{(a) A grasp pose in real execution. (b) The correspondent ideal grasp pose in simulation. (c.1) The error between the captured point cloud of the pen and the ideal pen pose. The point cloud is shown in green. The ideal pen pose is shown in red). (c.2) Comparison of the matched in-hand pose (red) to the ideal pose (yellow). (d) An example of a corrected drawing pose. The yellow robot configuration is corrected to the red configuration by the estimated in-hand pose.}
    \label{fig:refine}
\end{figure}

Fig.\ref{fig:refine}(a) and (b) show the grasp pose from pick-up motion in real execution and in simulation respectively. Fig.\ref{fig:refine}(c.1) shows the error between the captured point cloud of the pen and the pen pose in simulation. The point cloud is shown in green. The pen pose in the simulation is shown in yellow. Fig.\ref{fig:refine}(c.2) shows the matched pen pose with red color. The in-hand estimation corrects the pen pose from the yellow pose to the red pose. A transformation matrix for correction is computed as follows.
\begin{equation}
    _{Hand}^{Real}\textrm{T}=(_{Real}^{World}\textrm{T})^{-1}\times _{Sim}^{World}\textrm{T} \times _{Hand}^{Sim}\textrm{T}.
     \label{eq:trans}
\end{equation}
Here, $_{Hand}^{Real}\textrm{T}$ is the refined grasp pose and $_{Hand}^{Sim}\textrm{T}$ is the ideal grasp pose in simulation. Both of them are described in the pen's local coordinate system. Fig.\ref{fig:refine}(d) shows an example of a corrected drawing configuration. The yellow robot configuration is the one planned in simulation, while the red robot configuration is the corrected result. 

The red configuration in Fig.\ref{fig:refine}(d) is not necessarily solvable. When there is no solution, the planner invalidates the current grasp pose and trigger a new ``Grasp Pose Reasoning and Motion Planning'' routine, as is shown by the red arrows in Fig.\ref{fig:workflow}. The invalidation, re-reasoning, and re-planning close up the planning loop, thus improve robustness while maintaining high planning success rate.

\subsubsection{Hybrid Trajectory Following and Impedance Control}
\begin{figure}[!htbp]
    \centering
    \includegraphics[width=.825\linewidth]{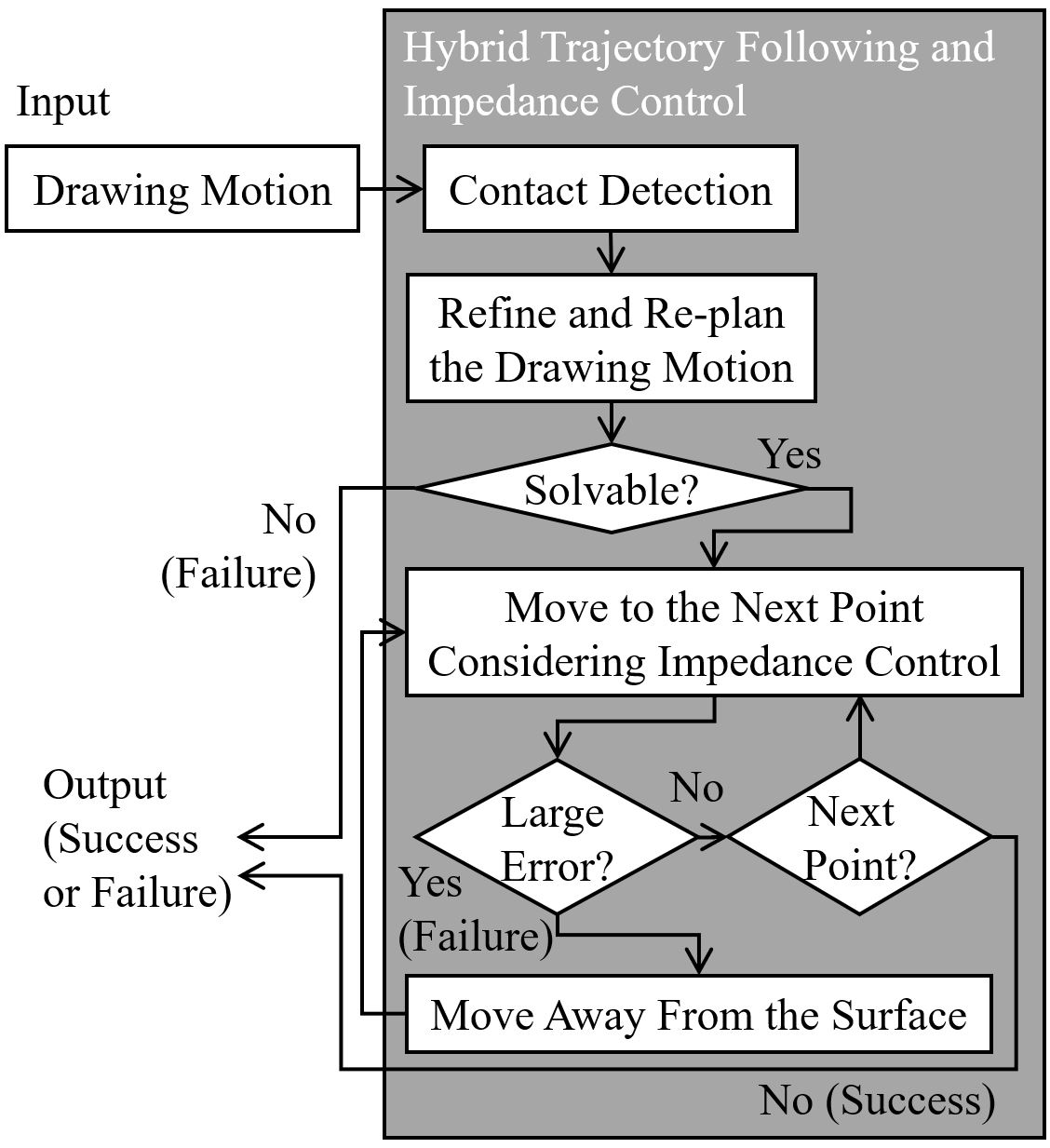}
    \caption{A detailed diagram of the ``Hybrid Trajectory Following and Impedance Control'' gray box in Fig.\ref{fig:workflow}.}
    \label{figimpedance}
\end{figure}

\paragraph{Impedance Control} We employ force feedback and impedance control to compensate for the errors caused by sensor calibration and the uncertainty in CAD models. Our impedance control method is originated from \cite{hogan1984impedance}. It builds a connection between the force and motion of the pen so that the robot can attach the pen to the target surface with flexible adjustment. Notably, we implement our impedance control by only considering compliance along the target surface's normal direction. As a result, the pen tip will follow the planned motion path while maintaining the firm contact with the target surface. Fig.\ref{figimpedance} shows the workflow of our impedance control component.

\paragraph{Error Detection and Recovery Considering Positional Deviation}
A drawback of incorporating impedance control is introducing new errors like (1) slippage of the pen on the target surface and (2) elastic changes in the shape of the target surface. We detect the failures by considering position feedback and close the loop up by ignoring the failed pen pose. If the deviation of the Tool Center Point (TCP) position is larger than a threshold, the motion will be paused. The pen will be pull up and move to the next pose to start a new iteration. The necessity of impedance control will be studied using comparison experiments in the experimental section.

\section{Experiments and Analysis}

This section presents the experiments carried out to analyze the robustness and flexibility of the developed robotic 3D drawing system. We use both simulations and real-world results to compare and verify the various methods. The simulation platform used in our experiments is a PC with Intel Core i5-8250U CPU and 32.00GB memory. The programming language is Python 3, and the software environment is WRS \footnote{https://gitlab.com/wanweiwei07/wrs} \cite{wan2016integrated}, which is an open-source environment developed in our laboratory. A Photoneo PhoXi 3D Scanner M with 772$\times$1032 resolution is used for visual detection. The real-world experiments are performed using a UR3 robotic arm, with a Robotiq Hand-E two-finger parallel gripper mounted at its end flange.

\subsection{Robustness of 2D-to-3D Stroke Mapping}
We evaluated the performance of proposed stroke mapping algorithms introduced in section IV with various 3D surfaces. Some of the surfaces have a known mesh model, while others do not. If the mesh model is not available, the ball pivoting algorithm \cite{bernardini1999ball}\cite{zhou2018open3d} is applied to reconstruct a mesh surface. 

For easy comparison, we specially present the visualized results of a lattice stroke. The lattice stroke is made of a series of boxes. Readers may easily estimate the performance of our mapping algorithms by watching the visualized results of mapped lattice strokes. Besides the visual estimation, we define two quantitative indicators to measure local and global deformations. The local indicator is
\begin{equation}
    local\_error = \frac{|\overrightarrow{q_{i}q_{i+1}}|-|\overrightarrow{p_{i}p_{i+1}}|}{|\overrightarrow{p_{i}p_{i+1}}|},
     \label{eq:error}
\end{equation}
where $p_{i}$ and $p_{i+1}$ are points in the stroke in $\mathbb{R}^2$. $q_{i}$ and $q_{i+1}$ are corresponding points in the mapped stroke in $\mathbb{R}^3$. The $local\_error$ shows the local variations after mapping while ignoring the necessary global bending.

The global indicator is
\begin{equation}
    global\_error = |\overrightarrow{q_{m}q_{n}}|-|\overrightarrow{p_{m}p_{n}}|
    \label{eq:gerror}
\end{equation}
where $p_{m}$ and $p_{n}$ are two closest points in different strokes in $\mathbb{R}^2$. $q_{m}$ and $q_{n}$ are their corresponding points in the mapped strokes in $\mathbb{R}^3$. For the lattice stroke, $|\overrightarrow{p_{m}p_{n}}|=0$.

\begin{figure*}
    \centering
    \includegraphics[width=.95\textwidth]{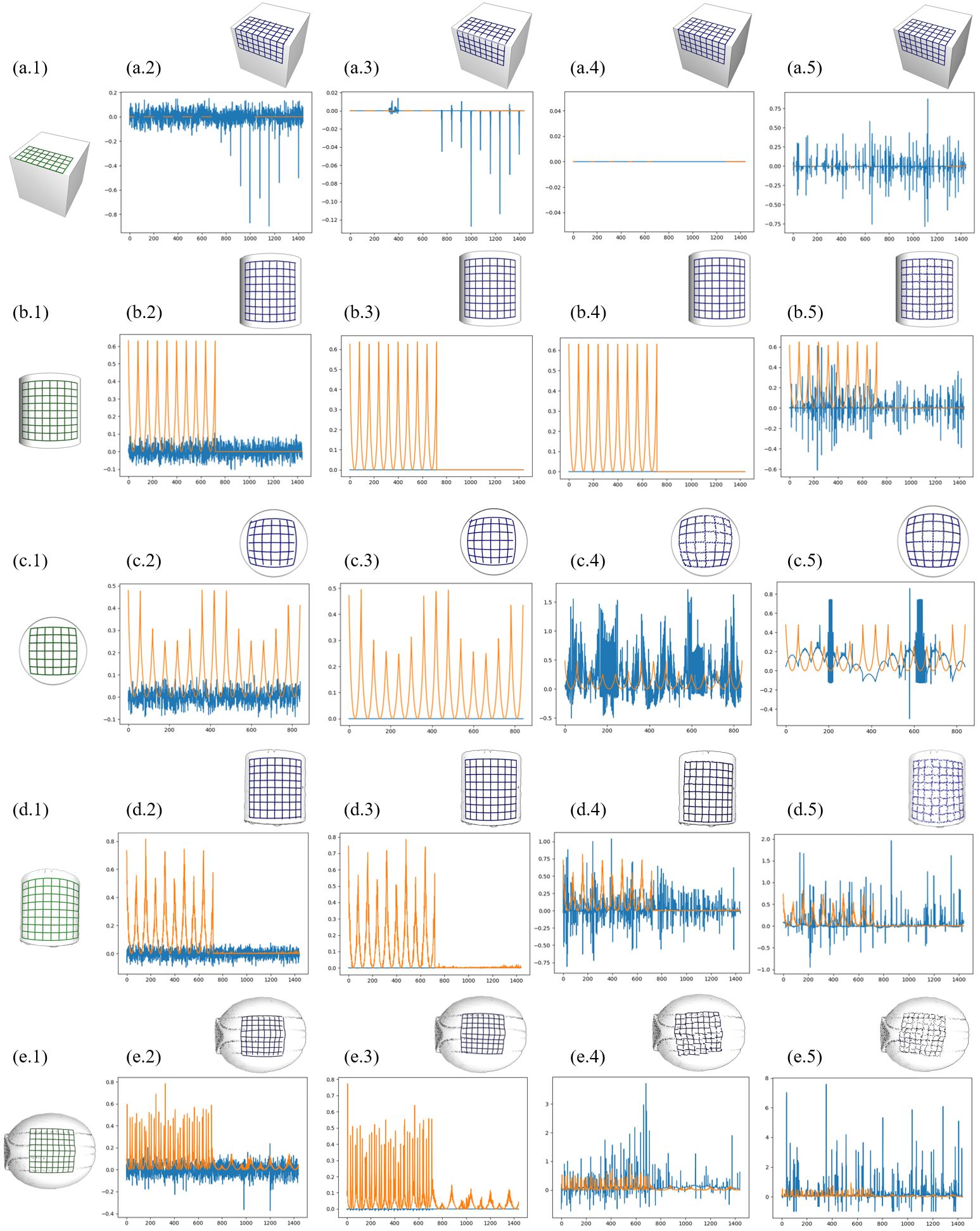}  
    \label{fig:projection_result_map}
    \caption{Results of 2D-to-3D stroke mapping. (a-c) Known mesh model. (d, e) Raw point clouds. (a.1-e.1) Results of the intuitive method. (a.2-e.2, a.3-e.3, a.4-e.4, a.5-e.5) Results of the DI, EI, SI, and II methods respectively. Each of the small subfigure identified by an (alphabet.number) in columns 2-5 includes two parts where the upper part is a visualization of the mapped result. The lower part is the error chart. The horizontal axis of an error chart is the point id. The vertical axis is the error value computed using equation \eqref{eq:error}. The blue curves in the error charts are the errors of the current method and model. The orange curve indicates the errors of the intuitive method. It is used as a baseline for comparison.}
    \label{fig:projection_result}
\end{figure*}

Table \ref{tab:projection_surface_info} shows the details about the various 3D surfaces and the lattice strokes used in our comparison studies. The table has two sections where the upper section shows the objects and surfaces with known mesh models. The lower section shows the ones without mesh models. In each section, there are seven rows where the first four are information about the target objects and surfaces, the fifth and sixth rows are the information about the lattice strokes, and the last row shows indices to the detailed mapping results in Fig.\ref{fig:projection_result}. 

The results in Fig.\ref{fig:projection_result} include both visualizations and local error comparisons using errors computed by equations \eqref{eq:error} an \eqref{eq:gerror}. The first column in Fig.\ref{fig:projection_result} and the orange curves in the error charts are the mapping results of the intuitive method mentioned in Section IV.A. It is considered as the baseline for comparison. The second to fifth columns of the figure shows the results of the DI, EI, SI, and II methods, respectively. Each of the small subfigure identified by an (alphabet.number) in the columns include two parts where the upper part is a visualization of the mapped result. The lower part is the error chart. The blue curves in the error charts are the local errors of the current method. The blue color is used here to have a significant difference with the orange color, so that readers may easily compare the current results with the baseline. The rows (a-c) are mapping results on surfaces with known mesh models. The rows (d,e) are the results with raw point clouds. One thing to note is that a closed mesh model should be decomposed into segments with shapes homographic to discs to realize the parameterization based methods (i.e., SI, II). We performed manual intervention to simplify the routine and reduce the error caused by the non-uniform parameterization scale between the segments. The mapping is carried out on the main segment that we selected.

\begin{table}[!htbp]
\renewcommand\arraystretch{1.2}
\centering
\caption{Information of the 3D Surfaces and Lattice Strokes}
    \begin{threeparttable}
    \begin{tabular}{@{\extracolsep{6pt}}cccc}
    \toprule
    & \multicolumn{3}{c}{Known Mesh Model} 
    \\ \cmidrule{2-4}
    & Box  & Cylinder   & Sphere       \\ 
    \midrule
    Volume (mm)      & 100$\times$100$\times$100 & 100$\times$100$\times$50     & 100$\times$100$\times$50   \\ 
    Surface (mm) & 80$\times$80   & 80$\times$80     & 60$\times$60 \\
    \# Vertices             & 6          & 360            & 32401      \\ 
    \# Faces                & 4         & 362            & 64440         \\ 
    \# Strokes              & 18             & 18             & 14  \\
    \# Points/Stroke              & 81     & 81     & 61  \\     
    Results  & Fig.\ref{fig:projection_result}(a) & Fig.\ref{fig:projection_result}(b) & Fig.\ref{fig:projection_result}(c) \\     
    \bottomrule 
    \toprule
    & \multicolumn{2}{c}{Raw Point Clouds} 
    \\ \cmidrule{2-3}
    & Cylinder & Helmet &  -   \\ 
    \midrule
    Volume (mm)      & 89$\times$101$\times$31 & 166$\times$199$\times$72  &  - \\ 
    Surface (mm) & 80$\times$80   & 80$\times$80     &  -\\
    \# Vertices & 12010 & 43617 &   -    \\ 
    \# Faces & 21407  & 80159 &     -     \\ 
    \# Strokes              & 18             & 18             & -  \\
    \# Points/Stroke           & 81    & 81    & - \\    
    Results  & Fig.\ref{fig:projection_result}(c) & Fig.\ref{fig:projection_result}(d) \\ 
    \bottomrule
    \end{tabular}
      \begin{tablenotes}
      \item[*] Meanings of items: Volume - The bounding box dimensions of a mesh model or point clouds; Surface - The 2D dimensions of the target surface for drawing; \# Vertices, \# Faces - The number of vertices and faces (original or reconstructed) on a target surface; \# Stroke - The number of strokes in a lattice to be drawn; \# Points/Strokes - The number of points in each stroke.
      \end{tablenotes}
    \end{threeparttable}
    \label{tab:projection_surface_info}
\end{table}


The results show that the metrology based mapping methods produce smooth strokes for both known mesh models and raw point clouds. Readers may compare Fig.\ref{fig:projection_result}(a.2-3, b.2-3, c.2-3) to Fig.\ref{fig:projection_result}(d.2-3, e.2-3) for better evaluation. The metrology-based methods tend to preserve local distances between adjacent points for the known mesh models, while the parameterization-based methods tend to preserve global relations. This point can be drawn by comparing Fig.\ref{fig:projection_result}(c.2-3) and (c.4-5). The strokes in Fig.\ref{fig:projection_result}(c.2-3) are less globally connected compared to those in Fig.\ref{fig:projection_result}(c.4-5). For the raw point clouds shown in the last two rows of Fig.\ref{fig:projection_result}, all methods have more distortions compared to mapping using known mesh models. 
The results also show that the performance of parameterization-based methods is unstable. For example, the SI method exhibits better performance on known mesh surfaces, as seen in Fig.\ref{fig:projection_result}(a.4, b.4). When the surface gets complicated, the distortion increases, as seen in Fig.\ref{fig:projection_result}(c.4). The reason is two-fold: (1) The parameterization scale is non-uniform. (2) The error of conformal mapping increased. Also, the parameterization-based methods will not work on more complicated models like models with concavity. Segmentation must be performed to make the methods applicable to the all models. Another observation is on the raw point clouds shown in Fig.\ref{fig:projection_result}(d.4-5, e.4-5), the mapping results get worse. The reason is also two-fold: (1) Controlling the parameterized results' rotation is difficult. (2) The data is noisy. 

\begin{table*}
\renewcommand\arraystretch{1.2}
\centering
\caption{Time Costs and Average Errors of the Mappings in Fig.\ref{fig:projection_result}}
    \begin{tabular}{@{\extracolsep{6pt}}ccccccccc}
    \toprule
    &  &  & \multirow{2}{*}{Baseline} & \multicolumn{3}{c}{Metrology Based Methods} & \multicolumn{2}{c}{Parameterization Based Methods}\\ 
    \cmidrule{5-7} \cmidrule{8-9}
    &  &  &  & DI & \textcolor{lightgray}{DI$^\prime$} & EI   & SI   & II\\
    \midrule
    \multirow{9}{*}{\begin{tabular}[c]{@{}c@{}c@{}}Known \\ Mesh Models\end{tabular}} 
    & \multirow{3}{*}{Box} 
    & Time Cost (s) & \textbf{2.040} & \colorbox{lightgray}{226.369} & \textcolor{lightgray}{1.331} & 37.643 & 2.759 & 36.640 \\
    &  & Avg. Local Error & - & 0.0055  & \textcolor{lightgray}{0.0147} & 0.0015 & \textbf{0.0} & \colorbox{lightgray}{0.0038} \\ 
    &  & Avg. Global Error & 0.0 & \colorbox{lightgray}{2.7591} & \textcolor{lightgray}{7.2060} & 0.1200 & 0.0 & 0.0 \\ 
    \cmidrule{2-9}
    & \multirow{3}{*}{Cylinder} 
    & Time Cost (s) & 7.195 & \textbf{2.206} & \textcolor{lightgray}{0.667} & 37.741 & 24.983 & \colorbox{lightgray}{68.732} \\
    &  & Avg. Local Error & 0.0795 & 0.0032 & \textcolor{lightgray}{0.0152} & 7.3580e-05 & \textbf{1.4870e-05} & \colorbox{lightgray}{0.0028} \\ 
    &  & Avg. Global Error & 0.0 & \colorbox{lightgray}{1.0983} & \textcolor{lightgray}{4.4346} & 0.0047 & 0.0 & 0.0 \\ 
    \cmidrule{2-9} 
    & \multirow{3}{*}{Sphere} 
    & Time Cost (s) & \colorbox{lightgray}{532.9310} & \textbf{2.394} & \textcolor{lightgray}{1.003} & 7.199 & 156.232 & 68.311 \\
    &  & Avg. Local Error & 0.0952 & 0.0046 & \textcolor{lightgray}{0.0881} &\textbf{0.0001} & \colorbox{lightgray}{0.2138} & 0.1176 \\ 
    &  & Avg. Global Error  & 0.0 & \colorbox{lightgray}{2.5859} & \textcolor{lightgray}{6.9255} & 2.4407 & 0.0 & 0.0 \\ 
    \midrule
    \multirow{6}{*}{\begin{tabular}[c]{@{}c@{}}Raw \\ Point Clouds\end{tabular}} 
    & \multirow{3}{*}{Cylinder} 
    & Time Cost (s) & \colorbox{lightgray}{310.930} & \textbf{2.122} & \textcolor{lightgray}{0.852} & 4.957 & 38.454 & 34.422 \\
    &  & Avg. Local Error & 0.0782 & 0.0033 & \textcolor{lightgray}{0.0467} & \textbf{7.4999e-05} &\colorbox{lightgray}{2.2308} & 0.0489 \\ 
    &  & Avg. Global Error & 0.0 & \colorbox{lightgray}{1.6907} & \textcolor{lightgray}{6.6618} & 0.3528 & 0.0 & 0.0 \\ 
    \cmidrule{2-9} 
    & \multirow{3}{*}{Helmet} 
    & Time Cost (s) & \colorbox{lightgray}{1154.746} & \textbf{2.843} & \textcolor{lightgray}{1.371} & 14.817 & 138.156 & 142.920 \\
    &  & Avg. Local Error & 0.0641 & 0.0090 & \textcolor{lightgray}{0.2038} & \textbf{0.0003} & \colorbox{lightgray}{2.0265} & 0.2313\\ 
    &  & Avg. Global Error & 0.0 & \colorbox{lightgray}{3.1040} & \textcolor{lightgray}{20.3439} & 2.0121 & 0.0 & 0.0 \\ 
    \bottomrule
    \end{tabular}
\label{tab:projection_result}
\end{table*}

Table \ref{tab:projection_result} shows the time costs and quantities of local and global errors of the proposed methods. The values with bold fonts show the minimum element of each row. The values with a light gray background are the maximum element of each row. By comparing the rows and columns, we find the metrology based methods are faster than the parameterization based methods. The average local errors of EI are less than 0.001. It has better precision compared to others. The DI method is the most lightweight one. It is much faster compared to others. Meanwhile, the precision of the DI method is still satisfying. However, the DI method has a significant drawback - Its precision depends on the density of the samples used to compute the nearest point to $q_{i+1}^{'}$. The data in the DI column of Table \ref{tab:projection_result} are the results using 1,000,000 samples in addition to the original vertices in the mesh model. When the number of samples decreases, the performance drops significantly. The DI$^\prime$ column of Table \ref{tab:projection_result} shows the results using 100,000 samples. Compared to the high-density column, both local and global errors increased significantly.

\begin{figure}[!htbp]
    \centering
    \includegraphics[width=.92\linewidth]{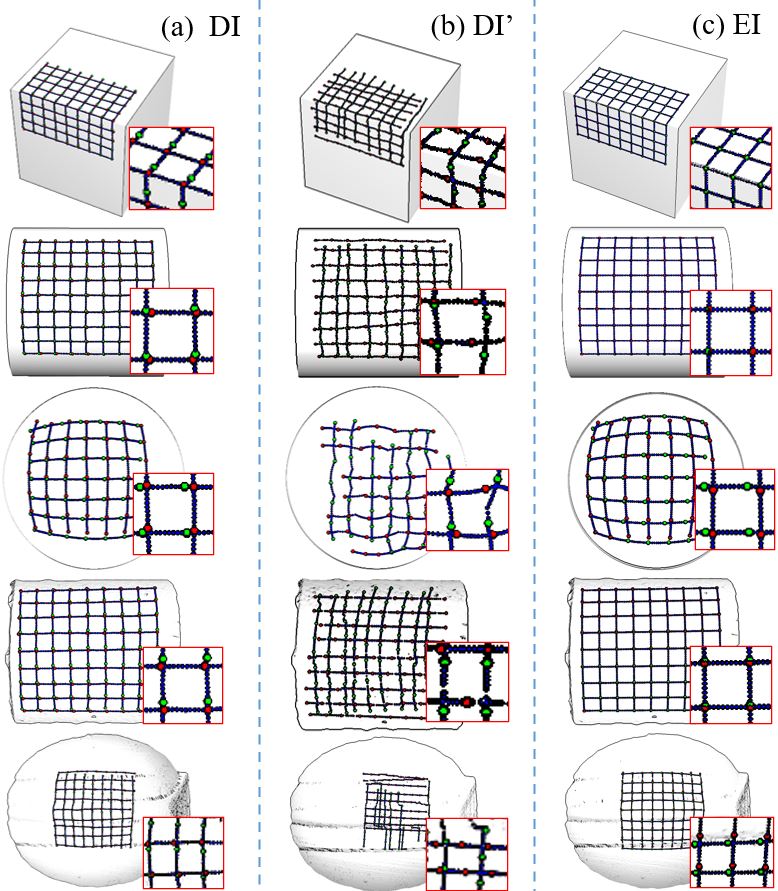}
    \caption{Results of the DI and EI methods with different sampling densities. (a) DI with 1,000,000 samples. (b) DI with 100,000 samples or less. (c) EI with the same number of samples as (b).}
    \label{fig:comparedensity}
\end{figure}

Fig.\ref{fig:comparedensity} visually compares in detail the results of the DI and EI methods with different sample density. With the same density, the EI method is much better than DI (the DI$^\prime$ column). It is even competitive with the results using 1,000,000 samples (the DI column). The red and green dots in Fig.\ref{fig:comparedensity} show the global error. They are the $q_{m}$ and $q_{n}$ of vertical strokes and horizontal strokes respectively. A larger displacement between the red and green dots indicate a higher global error. Compared to DI, EI has a better global performance. Note that as $p_{m}$ and $p_{n}$ in $\mathbb{R}^2$ have the same value, the baseline, SI, and II methods will map them without a difference. The global errors for these three methods are 0.

\subsection{Costs of Reasoning and Motion Planning}
\begin{figure}[!htbp]
    \centering
    \includegraphics[width=.87\linewidth]{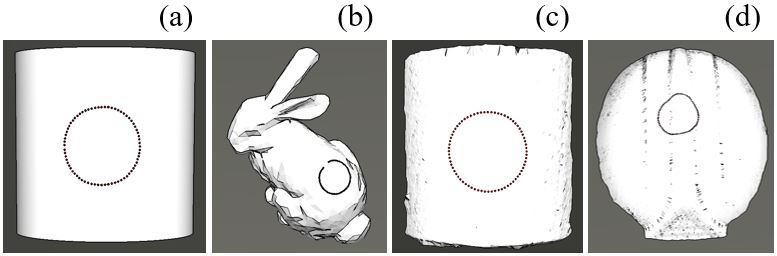}
    \caption{Four drawing tasks used to evaluate the performance and costs of the proposed reasoner and planner. The strokes are a same circle. (a, b) are mesh models. (a) is a cylinder. (b) is a Stanford bunny. (c, d) are point clouds. (c) is captured using the same cylinder as (a). (d) is captured using an engineering helmet.}
    \label{fig:planner_time_cost}
\end{figure}

We use the four drawing tasks shown in Fig.\ref{fig:planner_time_cost} to study our reasoner and motion planner's performance and costs. The strokes for the four tasks are the same -- a circle. It consists of 72 points. The first two target surfaces shown in Fig.\ref{fig:planner_time_cost}(a, b) are human-designed CAD models. There are known meshes for them. The target surfaces in Fig.\ref{fig:planner_time_cost}(c, d) are raw point clouds. The size of the mapped strokes varies according to the shape of the target surfaces.

The grasp reasoning and motion planning performance for the four tasks are summarized in Table \ref{tab:planner_time_cost}. The labels (a-d) in the first row indicates the correspondence of each column to the tasks shown in Fig.\ref{fig:planner_time_cost}. Below the first row, the table has three sections where the first section shows the size of the surfaces and the number of SLERPed pen poses along the mapped stroke. The following two sections show information about trials and failures in reasoning and the time costs. The results of the latter two sections are further analyzed in the following paragraphs. 

\begin{table*}
\renewcommand\arraystretch{1.2}
\centering
\caption{Costs of Grasp Reasoning and Motion Planning}
    \begin{tabular}{@{\extracolsep{6pt}}clcccc}
    \toprule
    \multicolumn{1}{l}{} & & (a) & (b) & (c) & (d) \\     
    \midrule
    \multirow{2}{*}{\begin{tabular}[c]{@{}c@{}} Information of the Surface and Strokes
    \end{tabular}} & Surface (mm) & 40$\times$40 & 25$\times$25 & 40$\times$40 & 40$\times$40 \\
    & \begin{tabular}[c]{@{}l@{}} \# SLERPed Pen Poses Along the Mapped Stroke\end{tabular} & 104 & 219 & 479 & 456\\ 
    \midrule
    \multirow{4}{*}{\begin{tabular}[c]{@{}c@{}} Grasps (62 Preannotated Grasp Poses in Total)\end{tabular}} & Index of 1st Common Grasp    & 1 & 12 & 1 & 0 \\
    & \# Common Grasps & 15 & 8 & 12 & 28 \\
    & Index of 1st Common Grasp with Solvable Motion & 1 & 12 & \colorbox{lightgray}{3} & \colorbox{lightgray}{4} \\
    & \# Common Grasp with Solvable Motion & 5 & 1 & 3 & 6\\ 
    \midrule
    \multirow{4}{*}{Time Costs} & Find the 1st Common Grasp (s) & 1.61 & 42.13 & 6.01 & 4.71\\
    & Find All Common Grasps (s) & 35.01 & 136.71 & 94.66 & 150.93\\
    & Find 1st Solvable Motion (s) & 38.54 & 12.70 & 28.35 & 44.14\\
    & Find All Solvable Motion (s) & 141.55 & 52.71 & 81.24 & 249.62\\
    \bottomrule
    \end{tabular}
\label{tab:planner_time_cost}
\end{table*}

We pre-annotated 62 candidate grasp poses for the pen. The reasoner iterated through the pen poses to find the common grasps for all key robot postures. The ``Grasps'' section of Table \ref{tab:planner_time_cost} includes four items where the first one shows the index of the 1st common grasp in the 62 pre-annotated grasps. The second item shows the number of all available common grasps. A motion planner will use a common grasp to generate the robot motion that moves between the adjacent key postures. The third item of the ``Grasps'' section shows the index of the 1st common grasp that has a solvable motion. For easy comparison with the first item, the index is also counted concerning the 62 pre-annotated grasps. Their values in (a) and (b) columns are the same as the first item, which means the first common grasp led to successful motion. Contrarily, the values in (c) and (d) are different from the first item, indicating that the planner encountered unsolvable motion and switched to different common grasps for replanning. The values in (c) and (d) are colored with a gray background to signify the observation. The fourth item of the ``Grasps'' section shows the number of all common grasps with solvable motion. Note that the total number of common grasps and common grasps with successful motion are counted for better analysis. They do not need to be fully scanned in practical applications. The reasoner and planner stop and the robot begins execution as long as the 1st successful motion is found. 

The ``Time Costs'' section shows the time needed to find the 1st common grasp, all common grasps, the 1st solvable motion, and all solvable motion. On average, the planning takes 30 $s$ to generate a drawing motion. The planning is more costly than reasoning.

\begin{figure}[!htbp]
    \centering
    \includegraphics[width=.97\linewidth]{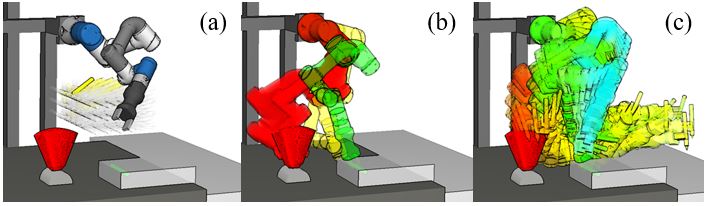}
    \caption{(a) Candidate pen poses. Red: The poses for drawing the mapped 3D strokes; Green: The detected initial pose; Gray: The intermediate poses for in-hand estimation. Yellow: The intermediate poses that share common grasps with the green and red poses. (b) One selected common grasp and its related IK solutions. (c) The successfully planned motion between adjacent IK solutions shown in (b).}
    \label{fig:keypose}
\end{figure}

Fig.\ref{fig:keypose} exemplifies one of the reasoned and planned results. The reasoner finds common grasps among the pen poses shown in Fig.\ref{fig:keypose}(a). Here, the red poses indicate the poses for drawing the mapped 3D strokes. The gray poses show the candidate positions for in-hand pose estimation. The green pose is the detected initial pose. The reasoner found common grasps among the red poses, the green poses, and the yellow-highlighted gray poses. Fig.\ref{fig:keypose}(b) shows one selected common grasp and its related IK solutions. The motion planner plans motion between adjacent IK solutions. A successful result is shown Fig.\ref{fig:keypose}(c).

\subsection{Performance of Real-world Executions}

We carry out robotic executions to study real-world performance. We specially concentrate on analyzing (1) the influence of in-hand pose estimation on the real trajectories, (2) the influence of impedance control on the drawing results, (3) the difference between model-based and point cloud-based drawing, (3) the flexibility for multiple pens and complicated strokes, and (4) error detection and recovery.

\subsubsection{The necessity of in-hand pose estimation}
\begin{figure}[!htbp]
    \centering
    \includegraphics[width=.97\linewidth]{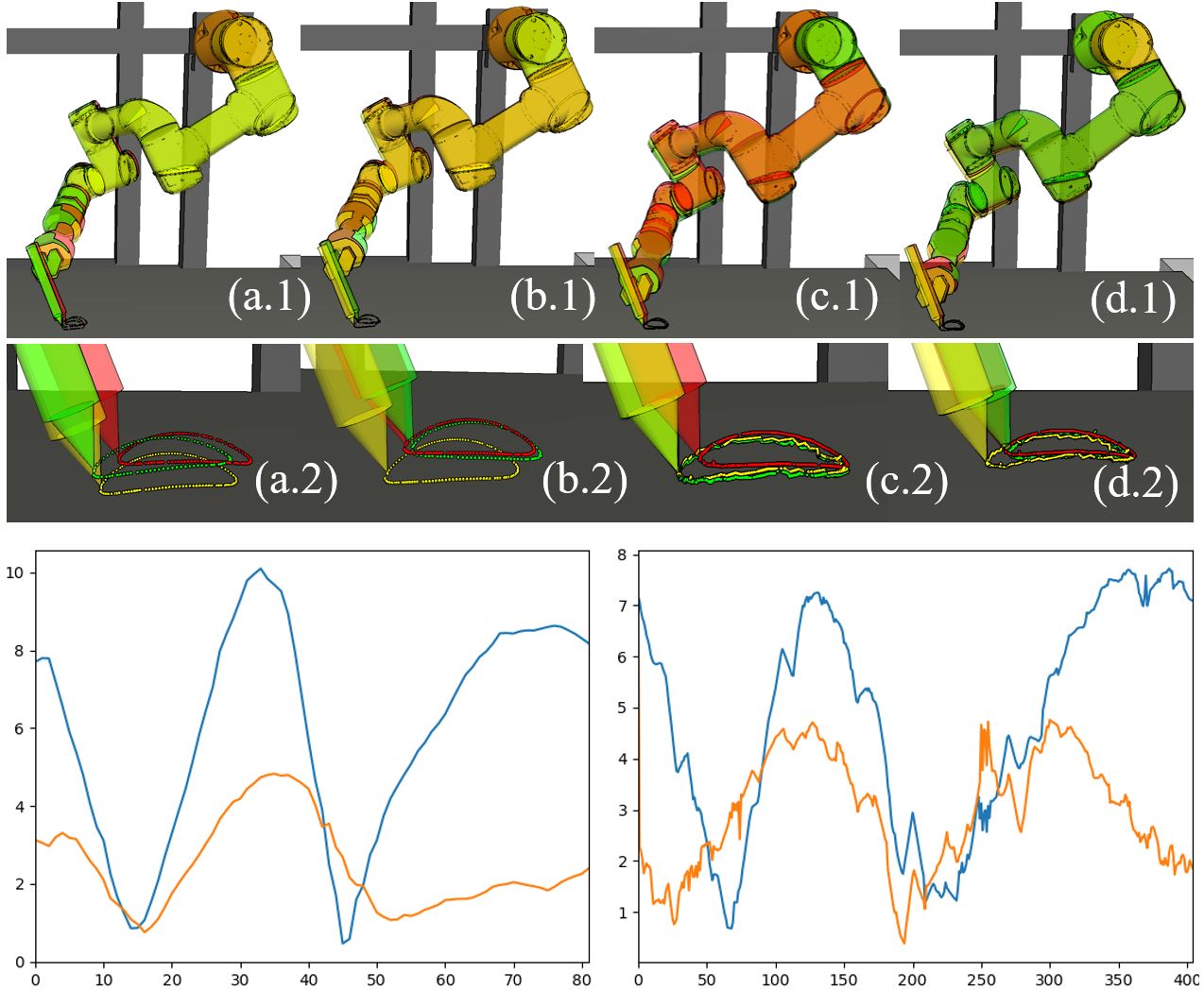}
    \caption{Comparison among the originally planned robot configurations (yellow robots and red strokes), the refined trajectories based on in-hand pose estimation (red robots and red strokes), and the real execution results using impedance control (green robots and green strokes). (a.1-2, c.1-2) The execution is performed using the originally planned trajectories. (b.1-2, d.1-2) The execution is performed using the refined trajectories. The results in (a, b) are based on the mesh model shown in Fig.\ref{fig:planner_time_cost}(a). The results in (c,d) are based on the point clouds shown in Fig.\ref{fig:planner_time_cost}(c). The bottom subfigures show the error curves. Blue: Original results; Orange: Results after in-hand pose estimation. Like Fig.\ref{fig:projection_result}, the horizontal axis is the point id. The vertical axis is the error values (difference between correspondent points).}
    \label{fig:inhand_estimation}
\end{figure}

We evaluated the necessity of the in-hand pose estimation by using the task of drawing a circle on a cylinder, like the ones seen in Fig.\ref{fig:planner_time_cost}(a, c). Both the mesh models and point clouds are used in the evaluation. Fig.\ref{fig:inhand_estimation} shows the results. The figure has two labeled rows (a.1-d.1) and (a.2-d.2). In the first labeled row, the yellow robot configurations indicate the planned results, while the red robot configurations indicate the refined motion after in-hand pose estimation. The green robot configuration is the real execution data read online from a working robot. The second row is a close-up view of the pen poses and pen tip trajectories for each subfigure in the first row. The different colors share the same meanings. 

The subfigures (a.1-2) and (b.1-2) are the results using a mesh model. The green configurations and trajectories in (a.1-2) are the originally planned motion's direct execution results. There was no in-hand pose estimation. Compared to the theoretical values (the red trajectories), there is a large offset. It indicates that direct executions suffer a lot from noise. The green trajectories in (b.1-2) are the execution result of the refined motion. Compared to the theoretical values, the offset is moderate. The results indicate that the in-hand pose estimation and refinement significantly improve precision. The subfigures (c.1-2) and (d.1-2) are the results using point clouds. Like (a.1-2) and (b.1-2), the two columns' green trajectories are the originally planned motion and refined motion execution results. The offset in these cases is smaller since the strokes are directly mapped to the measured surfaces, but there remains a discernible difference between the red and green trajectories.

The error charts after the two labeled rows in Fig.\ref{fig:planner_time_cost} visualize the distance between pen tip trajectory in simulation (red dot) and real execution (green dot). The horizontal axis of the figure is the point ids. The vertical axis is the difference between correspondent points. The blue curve is the difference before applying in-hand pose estimation. The orange curve is the result after in-hand pose estimation. The curves show that the offset decreased after in-hand pose estimation and motion refinement.

\subsubsection{The necessity of impedance control}
\begin{figure}[!htbp]
    \centering
    \includegraphics[width=.99\linewidth]{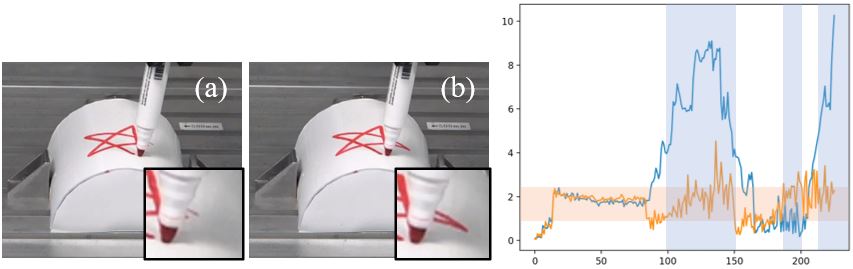}
    \caption{(a) Drawing a star without impedance control. (b) The result using impedance control. The chart shows the changes in forces. Horizontal axis: Time; Vertical axis: Magnitude of the force applied to the pen tip. The blue curve shows the changes in forces without impedance control. The orange curve shows the changes of force with force control.}
    \label{fig:comparison}
\end{figure}

The results in 1) partially show that impedance control is important to remove detection errors and ensure a solid drawing. In this part, we further examine the importance of impedance control by comparing the task shown in Fig.\ref{fig:comparison}. Here, the goal is to draw a star shape on a cylinder. Fig.\ref{fig:comparison}(a) shows the drawing result without impedance control. Fig.\ref{fig:comparison}(b) shows the drawing result with impedance control. Both methods can perform the drawing, but without impedance control, it is not easy to ensure the pen to be always in contact with the target surface. One tip of the star is lost when impedance control is not applied, as seen from the close-up boxes in the lower right corners of the two subfigures. The chart on the rightmost side of Fig.\ref{fig:comparison} compares the changes of forces during drawing. The horizontal axis is the point id. The vertical axis is the magnitude of the force applied to the pen tip. The blue curve shows the force changes without impedance control, while the orange curve shows the changes of force with force control. The curves imply that the force at the pen tip has large values between the 100th and 150th points and after the 220th point without force control. The force disappeared between the 180th and 200th points. We examined the real-world motion and found that the pen tip got lower than the surface when a large force appears (a heavy stroke), and got higher than the surface when the force disappears (a disappeared stroke). The surface used to generate the motion in the simulation environment was displaced. With the help of impedance control, the displacement error is successfully suppressed. The controller adjusts forces to be inside the orange range.

\subsubsection{Influence of mesh models}
\begin{figure}[!htbp]
    \centering
    \includegraphics[width=.93\linewidth]{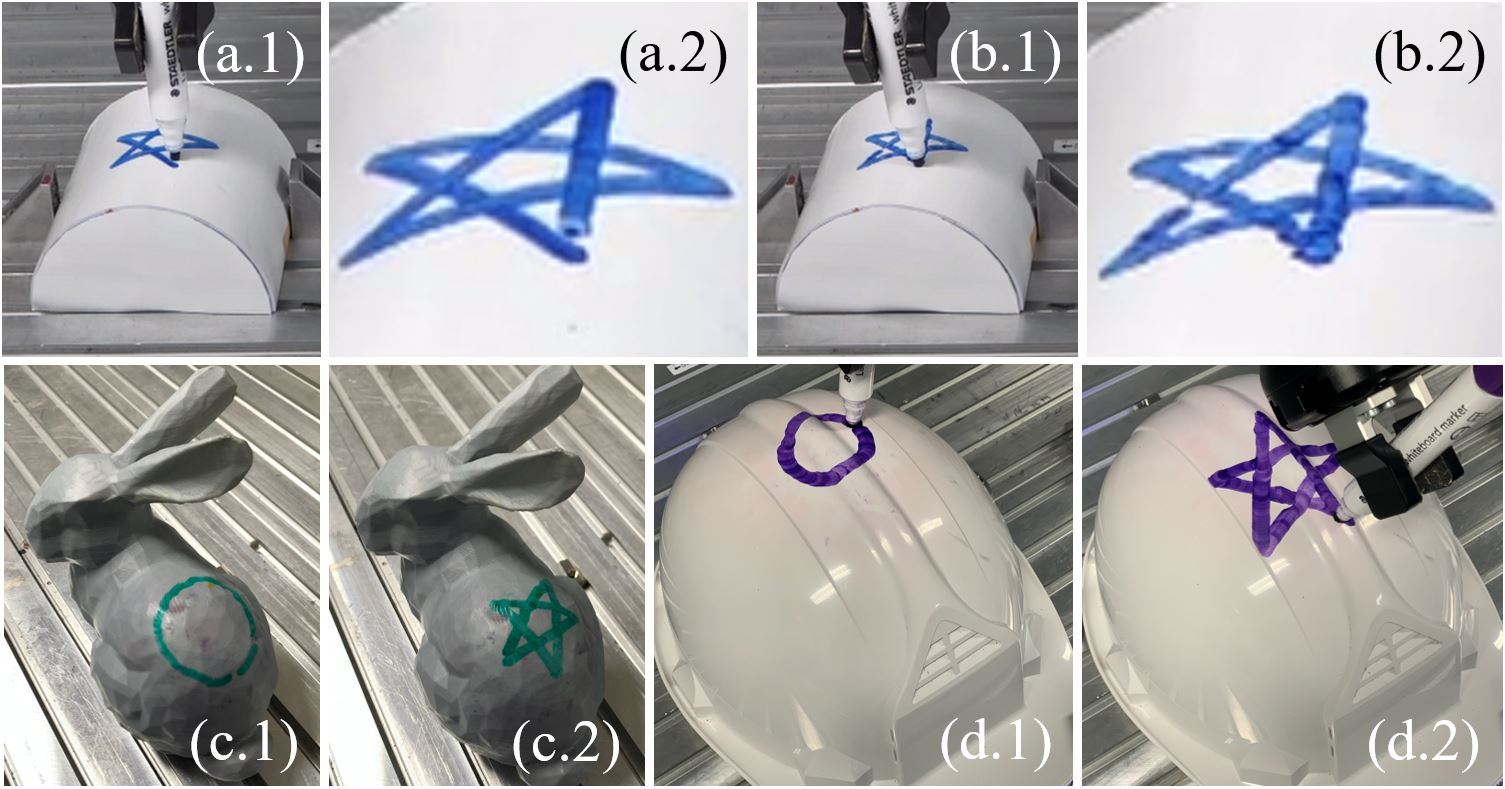}
    \caption{Comparing the drawing results planned using mesh models and point clouds. (a, b) Draw a star on the cylinder object. (a) The results with a mesh model. (b) The results using raw point clouds. (c, d) Other examples. (c) Drawing on a Stanford bunny. The mesh model is used. (d) Drawing on an engineering helmet. There is no mesh model.}
    \label{fig:bunny_helmet}
\end{figure}

We also compare the difference between the drawing results planned using mesh models and point clouds. The details are shown in Fig.\ref{fig:bunny_helmet}(a, b). The task is the same as the one shown in 2) - draw a star on the cylinder object. Fig.\ref{fig:bunny_helmet}(a.1) is the result based on a mesh model. Fig.\ref{fig:bunny_helmet}(a.2) shows a close-up view of the strokes. Fig.\ref{fig:bunny_helmet}(b.1) is the result based on point clouds. Fig.\ref{fig:bunny_helmet}(b.2) shows a similar close-up view. Our system can finish the drawing in both cases. However, the result planned with a mesh model is smoother than the result planned with point clouds, for the face normals estimated using point clouds are noisier than the inborn values of a mesh model. The pen poses of the model-based drawing are thus more continuous than those of the point cloud-based drawing. 

Fig.\ref{fig:bunny_helmet}(c, d) further show the results using two more difficult objects. Fig.\ref{fig:bunny_helmet}(c.1-2) show the results of drawing a circle and a star on the back of a Stanford bunny. The robot motion and pen trajectories are generated based on the bunny's mesh model. Fig.\ref{fig:bunny_helmet}(d.1-2) show the results of drawing the same circle and star on an engineering helmet. There is no mesh model for the helmet. The robot motion and pen trajectories are planned based on point clouds. Although the results based on point clouds are less smooth in Fig.\ref{fig:bunny_helmet}(d.1-2), they are only visible from a close-up viewpoint. From the viewpoints of Fig.\ref{fig:bunny_helmet}(c, d), the roughness is less noticeable and satisfying. Thus, we conclude that the developed robot system can draw on complicated surfaces both with and without mesh models. 

\subsubsection{Multiple pens, complicated strokes}

\begin{figure}[!htbp]
    \centering
    \includegraphics[width=.95\linewidth]{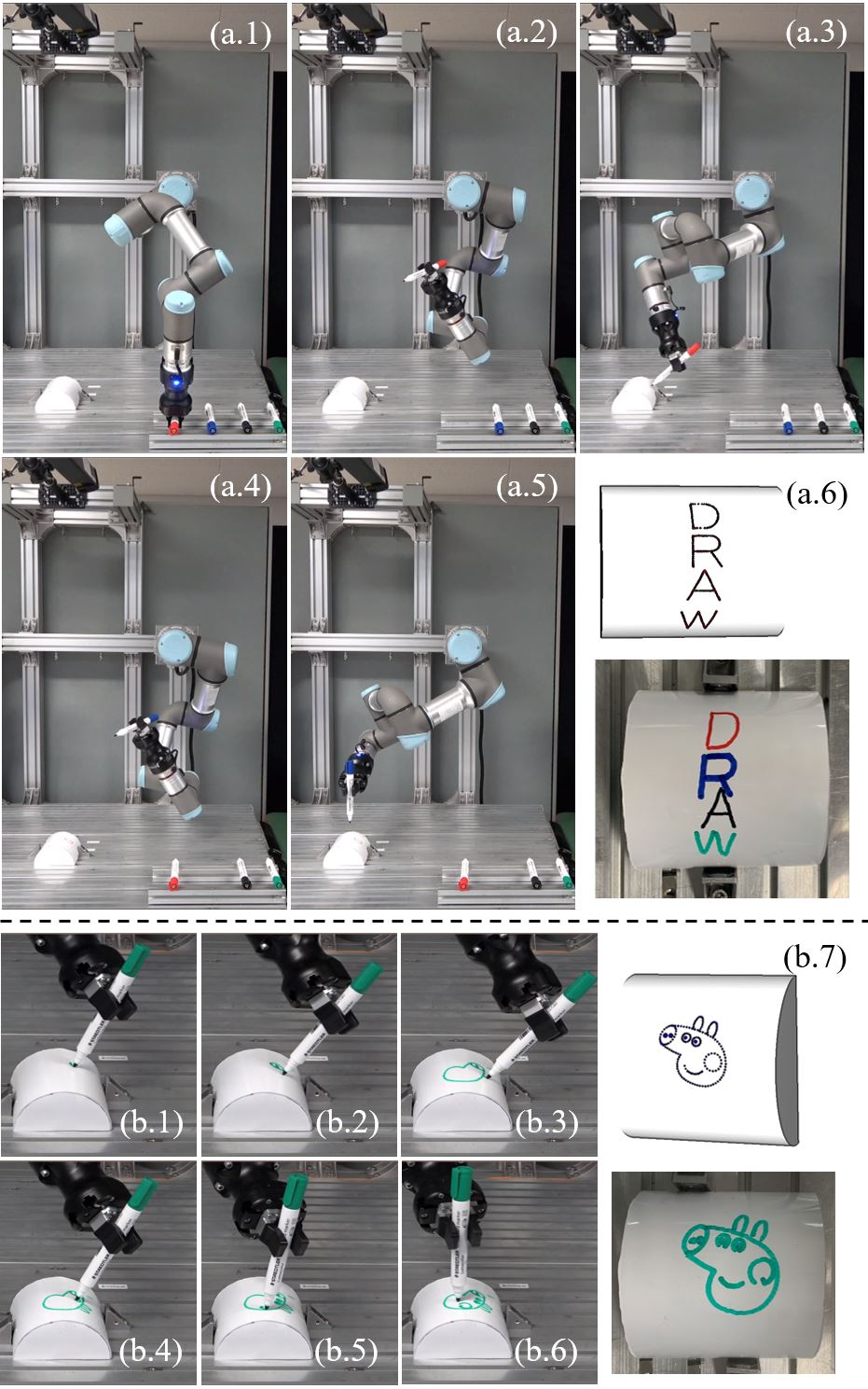}
    \caption{(a) Draw a word ``DRAW'' on a cylindrical surface with each letter in a different color. (b) Draw a Peppa Pig face on the same cylindrical surface.}
    \label{fig:multi}
\end{figure}

We examine the ability of our system to switch pens and draw complicated strokes using the two tasks shown in the upper part of Fig.\ref{fig:multi}(a.6) and (b.7). The first task is to draw the word ``DRAW'' on the cylindrical surface. Each letter in the word is required to have a specific color. The system successfully plans the motion for each pen as well as the motion for switching them. The Fig.\ref{fig:multi}(a.1-5) show snapshots of the execution sequence and the final drawing results. The second task is to draw a Peppa Pig face on the same cylindrical surface. The Peppa Pig face strokes include several small primitive elements. The developed system is still able to successfully and robustly perform the task. Fig.\ref{fig:multi}(b.1-6) shows snapshots of the execution sequence and the final drawing results of the Peppa Pig face.

\subsubsection{Error detection and recovery}
\begin{figure}[!htbp]
    \centering
    \includegraphics[width=.95\linewidth]{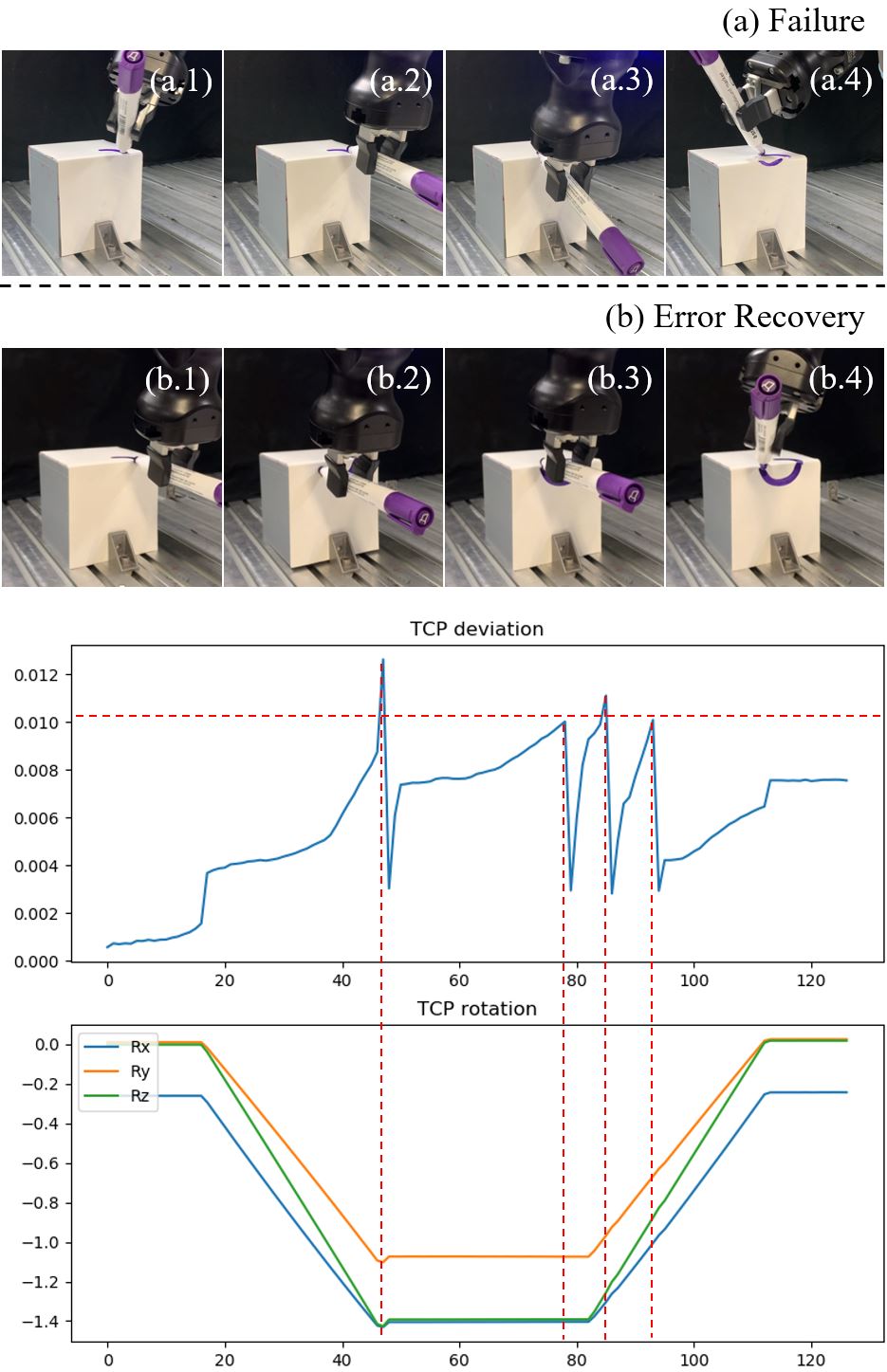}
    \caption{Draw a circle across the edge of a box. (a) The robot failed at a rectangular corner. There is no error detection and recovery. (b) The robots detected errors at the rectangular corners and recovered from the errors by moving ahead to the next point. Lower charts: The deviation in TCP positions and the corresponding TCP rotations. The horizontal axes are the point ids.}
    \label{fig:cube}
\end{figure}

Finally, we carry out experiments to study the error detection and recovery ability of the developed system. We designed a challenging task to test the ability. The goal is to draw a circle across the edge of a box, as shown in Fig.\ref{fig:cube}. Fig.\ref{fig:cube}(a) shows the results without error detection and recovery. In this case, the robot starts rotating the pen before reaching the edge. The following strokes become deviated and unpredictable due to continuous impedance control at the wrong positions. In contrast, Fig.\ref{fig:cube}(b) shows the results with error detection and recovery. The error detector compares the robot arm's TCP position to the refined values from in-hand pose estimation. If the error is larger than a given threshold, the robot will trigger recovery actions. The first chart in the lower part of Fig.\ref{fig:cube}(b) shows the deviation in TCP positions. At the points marked by the red dashed lines, the error detector finds large differences. At these points, the robot arm moves the pen away from the target surface, gives up the current point, and switches forward to the next point for recovering from the error. During the entire drawing on the box edge, the robot stopped four times and gave up four points, as seen in the TCP deviation chart. The second chart further compares the error points with hand rotation. The results show that three of the errors happen during rotation. They are caused by slippage at the pen tip. The other one is caused by the elastic changes in the target surface. The large pressing force resulted in the error. The proposed error detection and recovery method can deal with all these unexpected problems, enabling the developed robot system to draw the circle across the box's edge successfully.

\section{Conclusions and Future Work}
This paper presented a robotic drawing system that autonomously, robustly, and flexibly maps 2D strokes to 3D surfaces, plans pen manipulation motion, and performs drawing on 3D surfaces. The flexibility is ensured by treating pens as tools that can be picked up and manipulated. The robustness is achieved by high quality-stroke mapping, in-hand pose estimation and motion refinement, force control, and error detection and recovery. The system runs in a closed-loop manner to achieve maximum performance. Experimental results demonstrated the necessity of each craft. They together make the proposed system have an excellent expected performance. 

One limitation of the current system is that the input must be vectorized 2D strokes, which is difficult for non-professional users. In the future, we are interested in developing an interface that extracts strokes by watching a human demonstration.
\IEEEtriggeratref{34}
\bibliographystyle{IEEEtran}
\bibliography{references.bib}

\end{document}